\documentclass[letterpaper,twocolumn,10pt]{article}
\usepackage{usenix}
\usepackage{tikz}
\usepackage{filecontents}
\usepackage{xcolor}
\usepackage{amsthm}
\usepackage{booktabs}
\usepackage{times}
\usepackage{IEEEtrantools} 
\usepackage[numbers,sort&compress]{natbib}
\usepackage{caption}
\usepackage{tabularx}
\newcommand{\sysname}{\textsc{FedLeak}\xspace}
\newtheorem{thm}{\bf Theorem}
\usepackage{url}
\usepackage{algorithm}
\usepackage{algorithmicx}
\usepackage{algpseudocode}
\usepackage{listings}

\usepackage{graphicx}
\usepackage{amsmath} 
\usepackage{amssymb}
\usepackage{xspace}
\usepackage{multirow}
\usepackage{subfig}
\usepackage{pifont}
\usepackage[normalem]{ulem}
\usepackage{color}
\usepackage{marginnote}

\usepackage{enumitem}
\newcommand{\revise}[1]{\textcolor{black}{#1}}

\begin{document}

\date{}

\title{Boosting Gradient Leakage Attacks: Data Reconstruction in Realistic FL Settings}

\author{{\rm Mingyuan Fan$^1$} \hspace{5mm} {\rm Fuyi Wang$^2$} \hspace{5mm} {\rm Cen Chen$^{1}$}\thanks{Corresponding author.} \hspace{5mm} {\rm Jianying Zhou$^{3}$} \\
$^1$East China Normal University \hspace{1.5mm} $^2$RMIT University \hspace{1.5mm} $^3$Singapore University of Technology and Design \\
mingyuan\_fmy@stu.ecnu.edu.cn \hspace{5mm} fuyi.wang@rmit.edu.au \\
cenchen@dase.ecnu.edu.cn \hspace{5mm} jianying\_zhou@sutd.edu.sg }

\maketitle

\begin{abstract}
    Federated learning (FL) enables collaborative model training among multiple clients without the need to expose raw data.
    Its ability to safeguard privacy, at the heart of FL, has recently been a hot-button debate topic.
    To elaborate, several studies have introduced a type of attacks known as gradient leakage attacks (GLAs), which exploit the gradients shared during training to reconstruct clients' raw data. 
    On the flip side, some literature, however, contends no substantial privacy risk in practical FL environments due to the effectiveness of such GLAs being limited to overly relaxed conditions, such as small batch sizes and knowledge of clients' data distributions.
    
    This paper bridges this critical gap by empirically demonstrating that clients' data can still be effectively reconstructed, even within realistic FL environments.
    Upon revisiting GLAs, we recognize that their performance failures stem from their inability to handle the gradient matching problem.
    To alleviate the performance bottlenecks identified above, we develop \sysname, which introduces two novel techniques, partial gradient matching and gradient regularization.
    Moreover, to evaluate the performance of \sysname in real-world FL environments, we formulate a practical evaluation protocol grounded in a thorough review of extensive FL literature and industry practices.
    Under this protocol, \sysname can still achieve high-fidelity data reconstruction, thereby underscoring the significant vulnerability in FL systems and the urgent need for more effective defense methods.
\end{abstract}

\section{Introduction}
\label{sec_introduction}
Federated learning (FL)~\citep{fan2025trustworthiness,fed_survey} has become the de facto privacy-preserving approach for training deep neural networks (DNNs).
In FL, two steps are iteratively performed: 1) Each client downloads the current global model and computes gradients locally on its own data; 2) The server collects and aggregates these locally computed gradients from all participating clients to update the global model.
By transmitting only the gradients rather than raw data, FL inherently offers a plausible mechanism for privacy protection, promoting its adoption in various privacy-sensitive scenarios~\citep{common_batch,guardian}.

Although FL's privacy protection mechanism appears intuitive, recent studies have shown that clients' raw data can still be reconstructed by exploiting the uploaded gradients, known as gradient leakage attacks (GLAs)~\citep{dlg,invertinggrad,bayes_attack}.
We here focus on a passive attack scenario where the server does not manipulate the FL training process.
The core of GLAs lies in the gradient matching problem, which aligns the gradients generated from dummy data with the gradients uploaded from clients.
The vanilla GLA~\citep{dlg}, for example, starts by randomly initializing dummy data and labels, subsequently adjusting these initializations iteratively to approximate clients' raw data by minimizing the $L_2$ distance between the uploaded gradients and the gradients generated by these initializations.

However, the effectiveness of existing GLAs remains constrained to overly simplified scenarios, casting doubts about their actual threat in real-world FL environments.
\textit{\textbf{First}}, current GLAs~\cite{eval_frame1,eval_frame2,bayes_attack} work well only when facing small models, simplistic datasets, and limited batch sizes, which are rarely reflective of real-world scenarios.
While recent efforts have sought to extend these attacks to more sophisticated models and datasets, e.g., ResNet and ImageNet, the construction results remain either insufficient or contingent upon access to clients' batch normalization statistics~\cite{gradinversion} or their data distributions~\cite{grad_gene1,grad_gene2,grad_gene3}.
This reliance poses significant challenges, since clients are not obligated to share such statistics~\cite{fedbn}, and gathering large amounts of data similar to that of clients is rather difficult~\cite{sok_grad_leakage}, particularly in data-scarce domains like healthcare.
\textit{\textbf{Second}}, the performance of GLAs is positively correlated with the sensitivity of the input data to gradient changes.
High sensitivity allows attackers to more easily manipulate input data to generate gradients that align with the uploaded gradients.
Conversely, low sensitivity complicates the gradient matching problem, making it harder for attackers to identify how changes in input will affect the gradients.
Generally, this sensitivity peaks at the outset of training when the model weights are randomly initialized~\cite{bayes_attack,sok_grad_leakage}.
Yet, this sensitivity can be greatly reduced through the use of pre-trained weights, as pre-trained weights have already seen other data, making the resulting model less responsive to clients' data.
\textit{\textbf{Third}}, in practice, clients can adopt various defense strategies to mitigate the impact of GLAs~\cite{dp,soteria,guardian}.

To investigate whether FL can truly safeguard client privacy in real-world environments, we design \sysname.
To the best of our knowledge, \sysname represents the first attack method capable of reconstructing high-fidelity data from gradients shared by clients in practical environments.
Importantly, \sysname operates without the need for any extra resources, relying exclusively on the information allowed by the basic FL training protocol~\cite{common_batch}, such as the model and gradients.
This indicates that the attack implication of \sysname can be extended across almost all FL systems that adhere to the basic FL training protocol.
Moreover, we clarify that \sysname should not be misconstrued as a malicious threat to existing FL systems; rather, it is intended as a constructive resource aimed at deepening our understanding of FL's privacy protection mechanism and assessing the effectiveness of various defense methods.
Below we describe the challenges encountered in designing \sysname and present our key ideas in addressing these challenges.

\textit{\textbf{Challenge I: What limits the effectiveness of existing GLAs?}}
Instead of incrementally refining existing attack methods, we take a step back to re-examine the gradient matching problem itself to uncover the root causes behind the failure of current GLAs.
We start by analyzing whether the gradient matching problem is well-posed, a crucial factor in determining the feasibility of GLAs.
The analysis largely yields a positive response, challenging the common belief that the inherent multiplicity of solutions in the gradient matching problem leads to the suboptimal performance of current GLAs~\cite{grad_gene2,guardian}.
We argue that the underperformance of these attacks stems from their inability to properly solve the underlying gradient matching problem.
To bridge this gap, we derive a sufficient condition to address the gradient matching problem, providing a guiding principle for designing effective attack methods.

\textit{\textbf{Challenge II: How to implement the proposed guiding principle into attack methods?}}
Based on the proposed principle, we introduce two novel techniques, namely partial gradient matching and gradient regularization.
Partial gradient matching extends the original gradient matching problem by only matching a selected subset of the gradient elements, while gradient regularization penalizes the gradients of dummy data.
We demonstrate that both techniques help satisfy the derived sufficient conditions more effectively.
Moreover, we develop approximate solutions for these two techniques 
to address the prohibitively high computational cost associated with their original formulations, thereby enabling more efficient implementation in practical attack scenarios.

\textit{\textbf{Challenge III: How can we assess the attack performance of \sysname in real-world environments?}}
To evaluate the effectiveness of \sysname in real-world environments, it is essential to design a practical evaluation protocol.
To this end, we carefully review existing studies in GLAs, identifying eight key factors that significantly influence attack performance.
Next, we examine a wide range of FL studies and available industrial-grade FL libraries, particularly those in privacy-sensitive domains such as healthcare, so as to derive commonly used values for these factors.
Leveraging this knowledge, we align the identified influential factors with real-world environments to develop an evaluation protocol grounded in hands-on examples.
Through this protocol, we can gain a more practical implication of \sysname against FL's privacy protection ability in real-world environments.

\textbf{\textit{Contributions.}}
We highlight our contributions below:
\begin{itemize}[leftmargin=*,topsep=1pt] 
    \item Through our reexamination of the gradient matching problem, we pinpoint the root cause behind the failure of existing GLAs. We also derive a sufficient condition to resolve the gradient matching problem.
    \item We introduce \sysname, which involves two novel techniques: partial gradient matching and gradient regularization. We further present two approximate solutions to address their prohibitively high computational costs.
    \item We formulate a practical evaluation protocol and conduct extensive experiments to validate the attack performance of \sysname in accordance with this protocol. The attack results indicate that, even in real-world environments, FL still poses significant privacy leakage risks.
\end{itemize}

\section{Background \& Related Work}
\label{sec_related_work}
\subsection{Federated Learning}

We denote the global model as $F(\cdot, w)$ parameterized by $w$.
Let $U$ be the pool of available clients and $\mathcal{D}_i$ be the local dataset for the $i$-th client.
The training process unfolds through iterative rounds $t \in \{1, 2, \ldots, T\}$:
\begin{itemize}[leftmargin=*,topsep=1pt]
    \item \textbf{Model distribution and local training.}
    The server broadcasts the current model parameters $w$ to a subset of clients $U_t \subseteq U$ as participants for the $t$-th training round.
    Each client $i \in U_t$ subsequently samples a mini-batch $\mathcal{B}_i=(x_i, y_i) \subseteq \mathcal{D}_i$ to compute the local gradients $\nabla_{w} \mathcal{L}(F(x_i, w), y_i)$, where $\mathcal{L}$ denotes the loss function.
    \item \textbf{Gradient aggregation and model update.}
    The computed gradients $\nabla_{w} \mathcal{L}(F(x_i, w), y_i)$ are sent back to the server. The server aggregates these gradients, commonly by averaging them, and updates the global model.
\end{itemize}
Several challenges complicate the efficient implementation of FL.
Many clients possess limited computational power, memory, and bandwidth~\citep{iot}. 
To alleviate this, some studies~\cite{common_batch,fed_survey} suggested that selected clients update their local models multiple times, followed by sending the resulting model parameters to the server.
The server then averages these uploaded parameters to form a new global model.
However, this strategy risks causing local models to become overly specialized to their respective datasets, hindering the global model's peformance~\cite{fed_convergence}.
Notably, in this strategy, the server can deduce approximate gradients by comparing the parameters of the old global model and the newly uploaded model.
Another critical challenge arises from data heterogeneity across clients, known as the non-IID problem~\citep{common_batch,fedbn}, which can lead to slower convergence rates and performance degradation.
In this paper, we focus on GLAs.

\subsection{Gradient Leakage Attack}

Recent studies~\citep{dlg,idlg} have highlighted the potential of shared gradients to be exploited in data reconstruction.
Let $\hat{g}$ denote the gradients estimated by the server.
Deep leakage from gradient (DLG)~\citep{dlg} iteratively adjusts dummy data $x',y'$ to bring their generated gradients closer to $\hat{g}$, proving effective for small batches and simple networks:
\begin{equation}
\label{eq_1}
    x',y'=\arg \min_{x',y'} ||\nabla_{w}\mathcal{L}(F(x',w),y') -\hat{g}||_2^2.
\end{equation}
\citet{idlg} discovered that with small batches, exact ground-truth labels can be inferred by analyzing the signs of the shared gradients, enabling more efficient data reconstruction.
Subsequent works~\citep{label_infer1,label_infer2,label_infer3} developed approximate label inference techniques for larger batches.
\citet{invertinggrad} found that cosine distance is more effective than Euclidean distance for data reconstruction and introduced total variation to regularize the reconstructed images.

Recent advancements~\citep{gradinversion,grad_gene1,grad_gene2,grad_gene3} have improved GLAs by allowing the server to access more information from clients.
\citet{gradinversion} assumed that batch normalization statistics are accessible, which provide additional cues to regularize the reconstructed images.
\citet{bayes_attack} re-approached existing attacks through a Bayesian lens, proposing the Bayes attack to enhance reconstruction quality by leveraging prior knowledge.
Other studies~\citep{grad_gene1,grad_gene2,grad_gene3} took a step further by positing that servers possess knowledge of clients' data distributions, allowing the server to train a generative model for data reconstruction.
They optimized the latent vectors of the generative model to output images that can generate gradients similar to the shared gradients.
Nonetheless, some literature~\citep{sok_grad_leakage,eval_frame1,eval_frame2,out} has criticized these assumptions as not conforming to actual FL environments.

\subsection{Gradient Leakage Defense}

Defenses against gradient leakage can be classified into two main categories: cryptography-based methods~\citep{MPC1,MPC2,HE1} and perturbation-based methods~\citep{soteria,out,guardian}.
Cryptography-based methods secure gradients by ensuring that the server only sees the plaintext of the aggregated gradients~\citep{MPC1,MPC2,HE1,HE2}.
This makes data reconstruction more challenging because the aggregated gradients imply larger batches, expanding the search space and complicating reverse engineering.
However, cryptography-based methods rely on complex operations like modular arithmetic and exponentiation.
These result in significant computational and communication overheads, making them unsuitable for secure aggregation against gradient leakage in resource-constrained environments.
\revise{
Moreover, their efficacy diminishes if attackers can reconstruct high-fidelity data from larger batch gradients~\citep{soteria2}, and when the server is allowed to send maliciously-crafted parameters that can exclude non-interested clients' gradients from the decrypted gradients~\citep{model_inconsistency}.
}

Perturbation-based methods offer a more lightweight alternative by slightly modifying gradients to obscure the server.
This strategy, however, incurs a delicate trade-off between utility and privacy where more substantial modifications yield better protection but also degrade the gradient utility.
Differential privacy (DP)~\citep{dp} introduces random noises to perturb ground-truth gradients.
Gradient sparsification~\citep{grad_gene3} retains only the most significant gradient elements, and gradient quantization~\citep{grad_gene3} represents gradient values with fewer bits.
Recent advancements include Soteria~\citep{soteria,soteria2}, OUTPOST~\citep{out}, and Guardian~\citep{guardian}, which selectively perturb the most cost-effective gradient elements.

\begin{figure*}[!ht]
    \centering
    \subfloat[GD]{
        \includegraphics[width=0.237\linewidth]{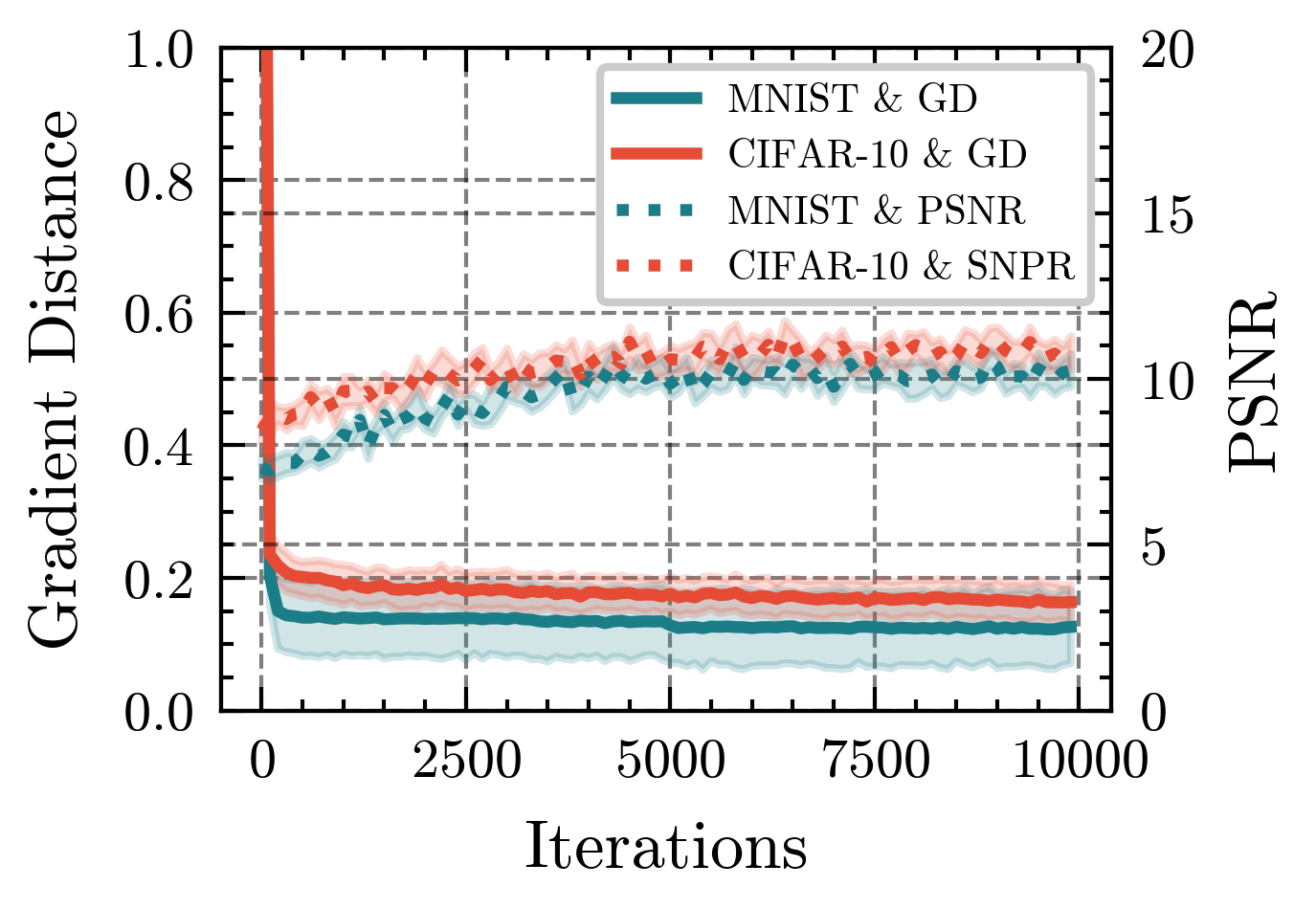}
        \label{fig_igla_loss_curve_sub2}
    }
    \subfloat[GD]{
        \includegraphics[width=0.240\linewidth]{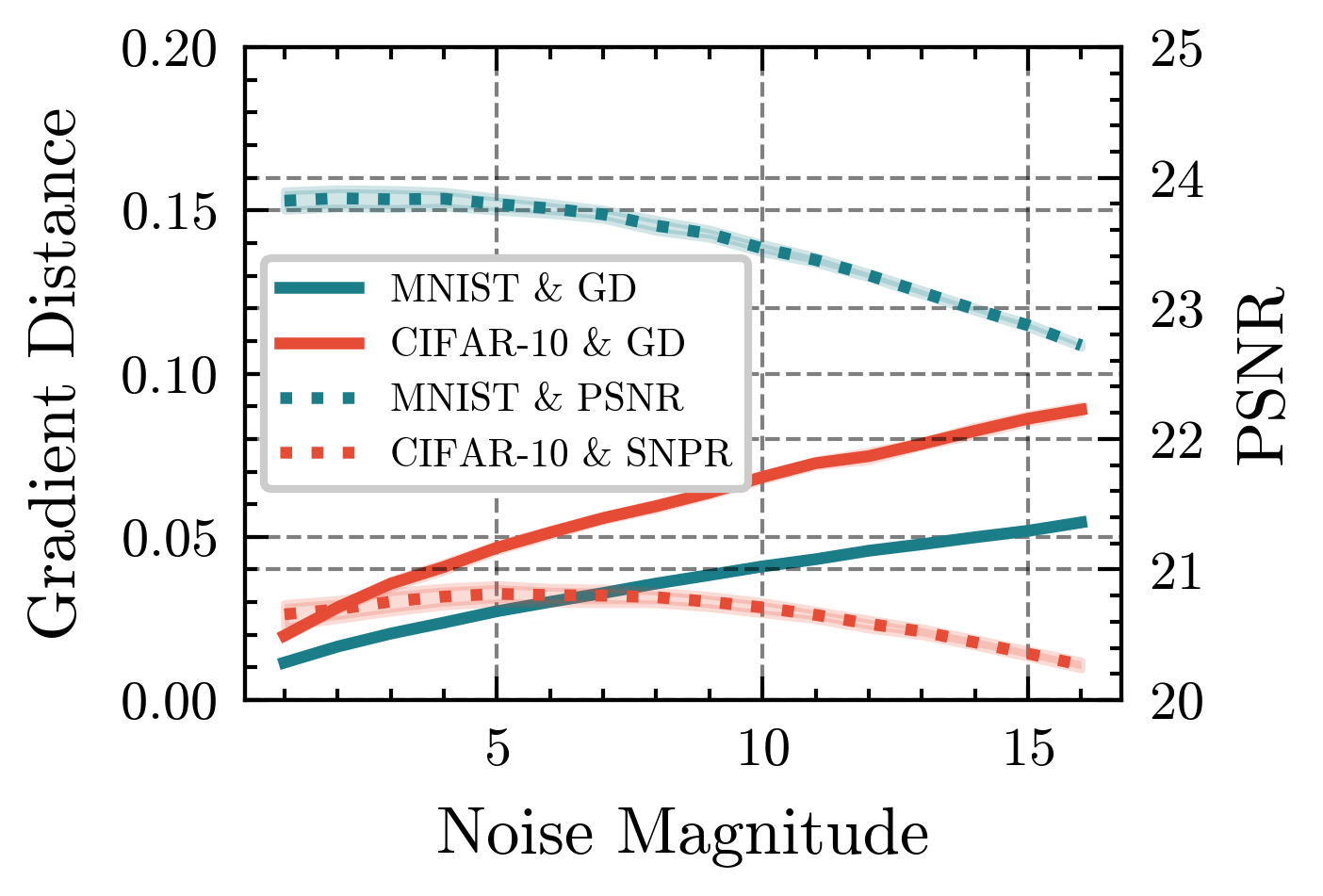}
        \label{fig_igla_loss_curve_sub3}
    }
    \subfloat[iDLG]{
        \includegraphics[width=0.222\linewidth]{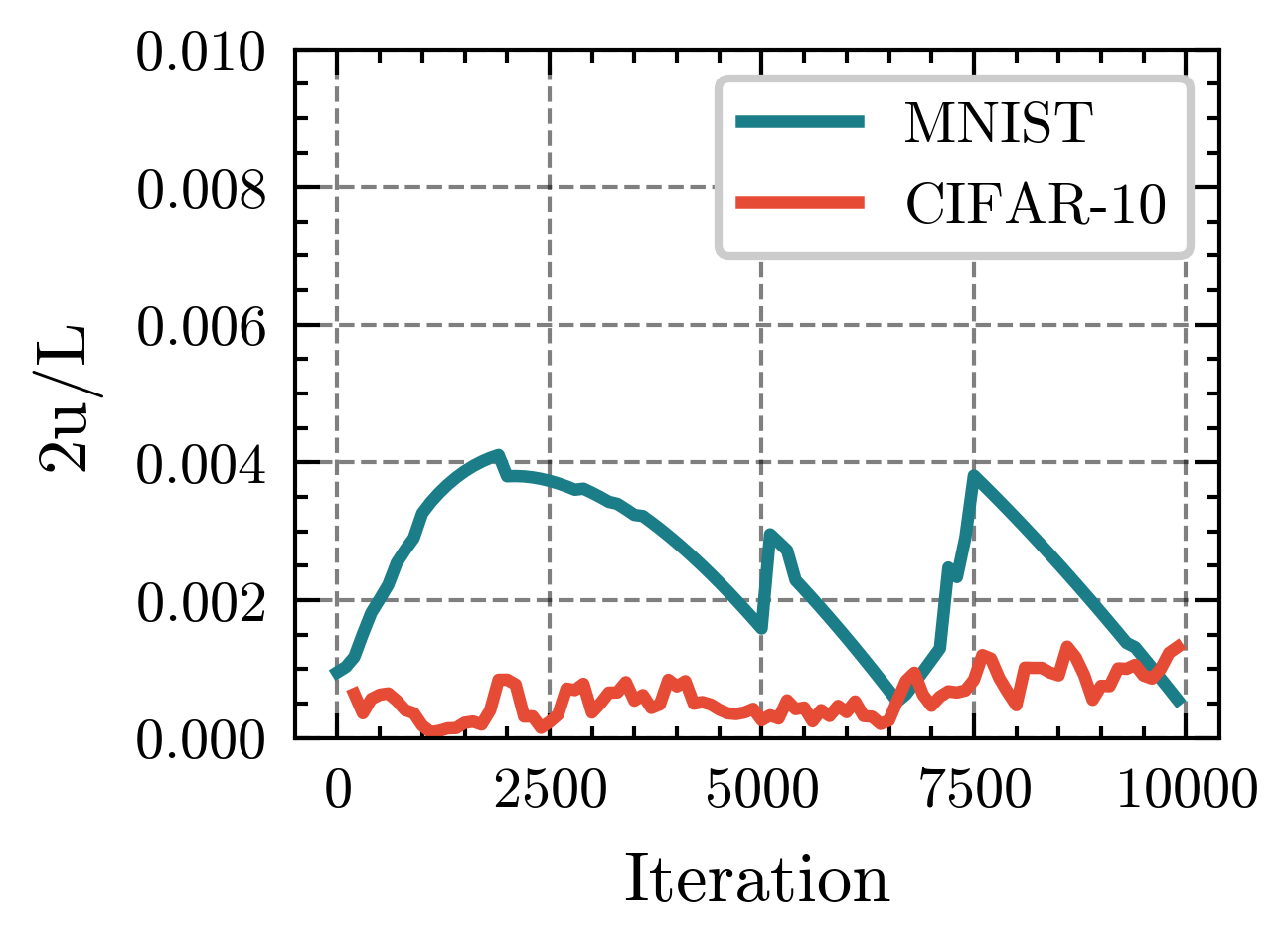}
        \label{fig_mu_l_sub1}
    }
    \subfloat[\sysname]{
        \includegraphics[width=0.214\linewidth]{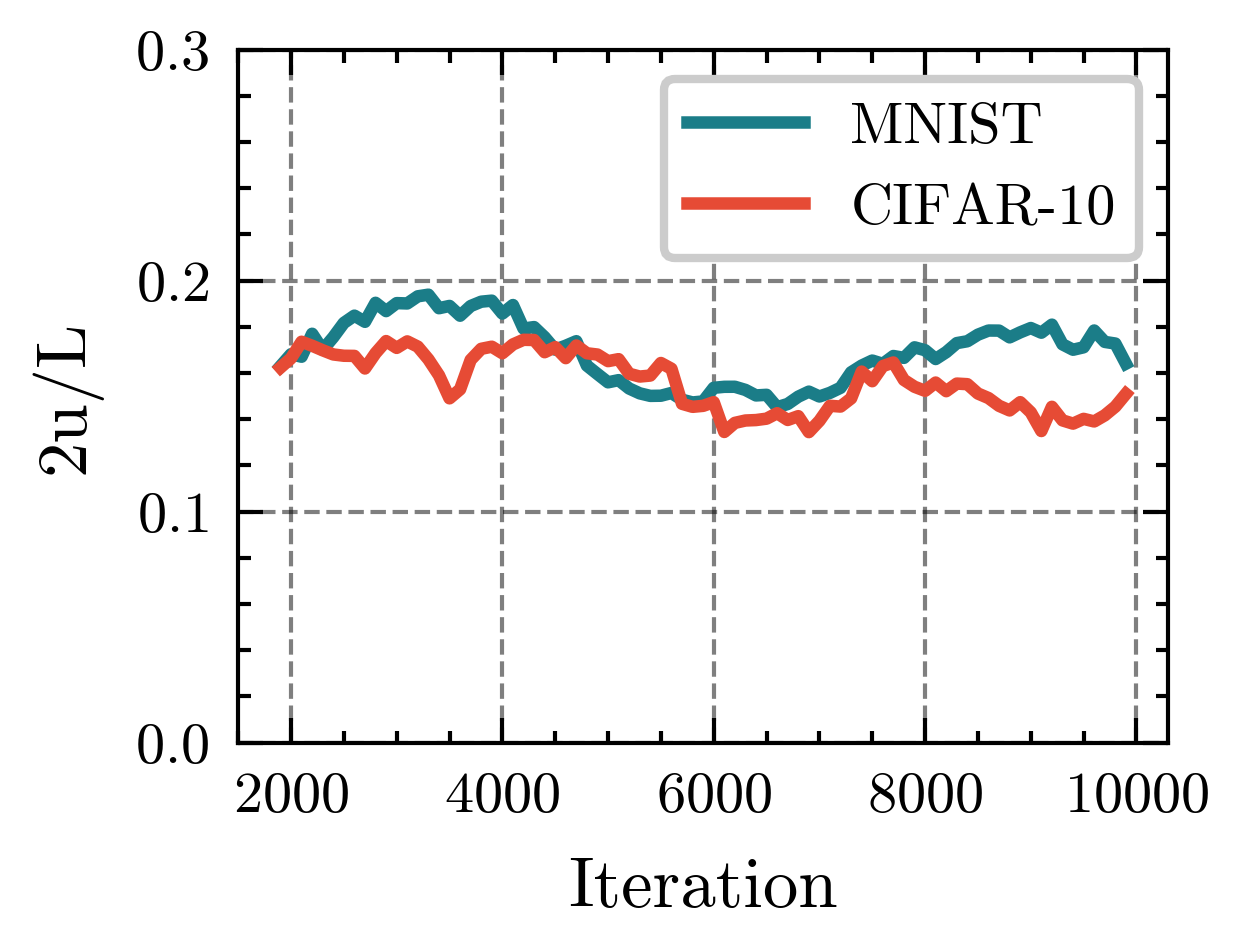}
        \label{fig_mu_l_sub2}
    }
    \caption{
    Figure~\ref{fig_igla_loss_curve_sub2} shows the $L_2$ GD between the images recovered by iDLG and the corresponding ground-truth images.
    Figure~\ref{fig_igla_loss_curve_sub3} demonstrates the $L_2$ GD between the ground-truth images and the same images with varying levels of random sign noise added.
    Figure~\ref{fig_mu_l_sub1} and Figure~\ref{fig_mu_l_sub2} illustrate the values of $\frac{2\mu}{L}$ associated with $x'$ achieved by the gradient descent algorithm and \sysname, both with a step size of $1 \times 10^{-4}$.
    All results are averaged over 100 samples.
    }
\end{figure*}

\subsection{Threat Model}
\label{sec_threat_model}

We consider the server to be honest-but-curious~\citep{dlg}, indicating that it honestly obeys the FL training protocol, yet attempts to recover clients' data from shared gradients.
This assumption is more reflective of real-world situations~\cite{common_batch,sok_grad_leakage} than a malicious model that permits the server to take malicious actions to facilitate reconstruction.
Such malicious actions, such as modifying the training procedure or model parameters~\citep{mali_server_attack1,mali_server_attack2,mali_server_attack3,mali_server_attack4}, could be easily spotted and undermine attack stealthiness~\cite{sok_grad_leakage}.
Moreover, given the effectiveness of label inference techniques~\citep{label_infer1,label_infer2,label_infer3}, the server's primary objective is to recover clients' data.

\textbf{Server's knowledge and capability.}
According to the basic FL training protocol~\cite{common_batch}, the server has full access to parameters of clients' uploaded models and the old global model, along with knowledge of training configurations such as learning rate and batch size.
The server is not allowed to access auxiliary information such as clients' batch normalization statistics and data distributions.
This restriction is justified, as practical clients are not required to furnish this information to the server \cite{sok_grad_leakage,eval_frame1,out}.
Notice that, while this limitation indeed increases our attack difficulty, it also broadens the applicability of \sysname across various FL scenarios.
Moreover, for thoroughness, we still include a comparative analysis of attack methods that assume access to this auxiliary information in our evaluations.
Finally, the server possesses moderate computational resources necessary for executing attacks.

We defer our evaluation protocol, i.e., the configuration of practical FL environments, to Section~\ref{sec_setup}.
For now, we just highlight that clients can perturb their ground-truth gradients to upload.

\section{Gradient Leakage Attack Revisited}
\label{sec_revisit}

For the sake of discussion, let us assume that the gradients $\hat{g}$ estimated by the server are identical to the gradients generated by local data $(x,y)$, i.e., $\hat{g}=\nabla_{w}\mathcal{L}(F(x, w), y)$. 
We also assume that the labels inferred by the server are the same as $y$.
Note that $\hat{g}$ can either be treated as the gradients from an individual client or as the aggregated gradients.
Let us first re-examine the gradient matching problem, formulated as follows:
\begin{equation}
\label{eq_2}
    x' = \arg \min_{x'} \text{dist}(\nabla_{w}\mathcal{L}(F(x', w), y), \hat{g}),
\end{equation}
where $\text{dist}(\cdot,\cdot)$ denotes a certain distance metric.

\textit{\textbf{The hypothesis of multiple solutions is likely a misunderstanding.}}
Most studies~\cite{idlg,eval_frame1,eval_frame2} have empirically demonstrated that images recovered through gradient optimization algorithms for Equation (\ref{eq_2}) are quite noisy, giving rise to the hypothesis of multiple noisy solutions beyond just $x$.
\revise{
This means different input samples can yield identical gradients\footnote{$f(x) = \sin(x)$ has the same gradients at $x = 0$ and $x = 2\pi$. In batches, gradient equivalence occurs through (1) direct gradient matching between distinct inputs, or (2) coincidental averaging of gradients from multiple samples (e.g., $f'(3) + f'(-1) = f'(0) + f'(2)$ for $f(x) = x^2$ where the gradients at any two points are not the same). We here examine the former one.}.
}
To illustrate this rmore concretely, consider the case of training a three-hidden-layer network with ReLU activation function on two benchmark datasets MNIST and CIFAR-10.
We employ iDLG~\citep{idlg} and observe the reconstruction performance of this attack with a batch size of 1 (see Appendix~\ref{appendix_exp_section3} for more attacks).
Figure~\ref{fig_igla_loss_curve_sub2} illustrates that the $L_2$ gradient distance (GD) and PSNR between $x'$ and $x$ appear to converge to approximately 0.15$\sim$0.2 and 10$\sim$11, respectively.
A PSNR value of 10$\sim$11 indicates that the recovered images are notably noisy~\cite{guardian}.
Furthermore, the $L_2$ GD of 0.15$\sim$0.2 seems quite small at first glance\footnote{The absolute difference between two gradient elements is on the order of $10^{-4}$.}, potentially leading to an interpretation that GLAs find noisy data that produce gradients almost identical to those of $x$, thereby reinforcing the hypothesis of multiple solutions.
We first clarify that this interpretation is a misunderstanding.
Specifically, the recovered noisy images do not constitute a solution to Equation (\ref{eq_2}), rather, they are attributable to an improper handling of Equation (\ref{eq_2}).

To substantiate the above claim, we generate Gaussian random noises for 100 input images, then multiply its sign by a strength of $\frac{1}{255} \sim \frac{16}{255}$, and add them to the input images.
This process is repeated 100 times, and Figure~\ref{fig_igla_loss_curve_sub3} shows the expected $L_2$ GD for varying noise strengths.
Empirical studies~\cite{adv_robust} say that, for low-resolution images ($32 \times 32$), the $L_{\infty}$ distance between two similar images should be less than $\frac{8}{255}$.
Generally, similar gradients arise from similar images, suggesting that the threshold for assessing gradient similarity can be roughly approximated by the GD between an image and its perturbed counterpart at a noise strength of $\frac{8}{255}$.
However, as shown in Figure~\ref{fig_igla_loss_curve_sub3}, even when the noise strength reaches as high as $\frac{16}{255}$, the expected $L_2$ GD stays below 0.1.
Consequently, a loss value (GD) around 0.2, as achieved by iDLG, signifies fundamental non-convergence in Equation (\ref{eq_2}) and falls far short of indicating that the dummy data have similar gradients to $x$.
Besides, this non-convergence is unlikely due to being trapped in local optima or saddle points, as the use of multiple restarts~\cite{pgd} and temporarily large learning rates~\cite{cos_lr} did not ameliorate the situation (Appendix~\ref{appendix_exp_section3}).
To further clarify this point, we now take a closer look at Equation (\ref{eq_2}).

\textit{\textbf{The examination of Equation (\ref{eq_2}).}}
At its core, Equation (\ref{eq_2}) can boil down to solving the following system of nonlinear equations:
\begin{equation}
\label{eq_3}
\nabla_{w}\mathcal{L}(F(x', w), y)[i]=\hat{g}[i], \ i=1,\cdots,|\hat{g}|.
\end{equation}
Due to $\hat{g}=\nabla_{w}\mathcal{L}(F(x, w), y)$, the system of nonlinear equations has at least one solution $x'=x$, suggesting that the system is not ill-conditioned.
Furthermore, it is highly likely that the solution is unique, i.e., $\nabla_{w}\mathcal{L}(F(x', w), y) \neq \nabla_{w}\mathcal{L}$ $(F(x, w), y)$ for $x' \neq x$, for several reasons.
Firstly, in Equation (\ref{eq_2}), the number of constraints (the number of model parameters) significantly exceeds the number of variables (the dimensionality of the input data)\footnote{For example, a ResNet10 model for MNIST has 4902090 parameters. For batch sizes of 1, 8, 32, 64, and 128, the corresponding input dimensions are 784, 6272, 25088, 50176, and 100352 respectively. The number of model parameters is 6252, 781, 195, 97, and 48 times the input dimensions for these batch sizes respectively.}. 
The over-constrained nature of the system reduces the chances of multiple solutions existing.
Actually, for fully connected neural networks, the solution is guaranteed to be unique, as substantiated by Theorem~\ref{thm1}.
Moreover, we highlight that Theorem~\ref{thm1} can, to some extent, be applied to convolutional neural networks (CNNs) and transformers, as convolutional layers and attention layers can be viewed as specialized variants of fully connected layers~\cite{training_survey}.

\vspace{+0.1cm}
\begin{thm}
\label{thm1}
     \textit{Consider $F$ whose first hidden layer is a fully connected layer, followed by at least one more fully connected layer. If gradients of $\mathcal{L}$ with respect to the outputs of these fully connected layers are not the zero vectors, then for any $x' \neq x$, it holds that $\nabla_{w}\mathcal{L}(F(x', w), y) \neq \nabla_{w}\mathcal{L}(F(x, w), y)$~\citep{invertinggrad}.}
\end{thm}
\vspace{+0.1cm}

Intuitively, different inputs generally yield distinct intermediate outputs within $F$.
These intermediate outputs affect the computation of the gradient of $F$'s parameters, making it challenging for different inputs to produce identical gradients.
In other words, if there exist two different inputs that generate identical gradients, it would imply that $F$’s error feedback in the parameter space for these inputs is the same, which is highly improbable unless the network degenerates significantly, for example, if all parameters are zero.
In summary, the noisy solutions are primarily due to the ineffectiveness of existing optimization algorithms in solving Equation (\ref{eq_2}), motivating us to identify the specific conditions that allow for effective solutions to Equation (\ref{eq_2}).

\textit{\textbf{Convergence conditions.}}
Let $g'=\nabla_{w}\mathcal{L}(F(x', w), y)$.
We use $x_j'$ and $g_j'$ to denote the starting point for $x'$ and the gradients generated by $x_j'$ at the $j$-th iteration.
The update rule of the gradient descent algorithm for $x'$ at $j$-th iteration is given by $x_{j+1}' = x_{j}' - \eta \nabla_{x_{j}'} \text{dist}(g_j', \hat{g}) $ where $\eta$ is the step size.
To ensure convergence of $x'$ toward $x$, we require $||x-x_{j+1}'||_2^2 < ||x-x_{j}'||_2^2$.
Then, expanding $||x-x_{j+1}'||_2^2$ produces:
\begin{equation}
\label{eq_4}
\begin{split}
    &||x-x_{j+1}'||_2^2 = ||x-x_{j}' + \eta \nabla_{x_{j}'} \text{dist}(g_j', \hat{g})||_2^2 \\
    &= ||x-x_{j}'||_2^2 - 2 \eta \langle x_j' - x, \nabla_{x_{j}'} \text{dist}(g_j', \hat{g}) \rangle\\
    &+ \eta^2 ||\nabla_{x_{j}'} \text{dist}(g_j', \hat{g})||_2^2.
\end{split}
\end{equation}
Let us impose the following conditions to hold:
\begin{equation}
\label{eq_5}
\begin{split}
    \langle x_j' - x, \nabla_{x_{j}'} \text{dist}(g_j', \hat{g}) \rangle &> \mu \ ||x - x_{j}'||_2^2, \\
    ||\nabla_{x_{j}'} \text{dist}(g_j', \hat{g})||_2^2 &< L \ ||x - x_{j}'||_2^2.
\end{split}
\end{equation}
Substituting the above conditions into Equation (\ref{eq_4}) yields:
\begin{equation}
\label{eq_6}
    ||x-x_{j+1}'||_2^2 < (1+ \eta^2 L -2 \eta \mu) ||x-x_{j}'||_2^2.
\end{equation}
To guarantee convergence, it is essential that $1+ \eta^2 L -2\eta \mu \leq 1$, i.e., $0 < \eta < \frac{2 \mu}{L}$.
Here, $\mu$ measures the curvature of the loss landscape near minima, while $L$ bounds the gradient smoothness (smaller $L$ implies smoother, more consistent gradients).
Larger $\mu$ and smaller $L$ together broaden the range of safe learning rates $\eta$ that ensure convergence.
We further evaluate the value of $\frac{2\mu}{L}$ associated with $x'$ over different optimization iterations, as shown in Figure~\ref{fig_mu_l_sub1}.
We see that the value of $\frac{2\mu}{L}$ is very small and susceptible to fluctuations, which poses a significant challenge in determining an appropriate step size $\eta$ for addressing the gradient matching problem with gradient optimization algorithms.
This also demonstrates that the failure of existing GLAs is due to a lack of an effective way to solve the gradient matching problem, rather than the existence of multiple feasible solutions.
In light of these observations, we develop \sysname, designed to more effectively resolve Equation (\ref{eq_2}).
The details of \sysname will be elaborated in the following section.

\textbf{Remark.}
The analysis in this section relies on the idealized assumption of $\hat{g} = \nabla_{w}\mathcal{L}(F(x, w), y)$, which may not hold.
This is primarily because clients might perform multiple local steps or upload perturbed gradients, which obfuscates the server from obtaining the exact gradients with respect to $(x,y)$.
Despite this difference from reality, we highlight that, in most cases, $\hat{g}$ are still close enough to $\nabla_{w}\mathcal{L}(F(x, w), y)$ to yield meaningful reconstructions (as shown in Section~\ref{eval_diverse_settings}), since excessive perturbation or an overly large number of local steps can lead to a significant degradation in model performance.
Additionally, we provide an analysis of the impact of such estimation errors in Appendix \ref{appendix_grad_est_error} for interested readers seeking further insights.
We will not elaborate on this here, as it is not the primary focus of this paper.

\section{Our Attack: \sysname}
\label{sec_approach}

\begin{figure*}[!ht]
    \centering
    \includegraphics[width=1.0\linewidth]{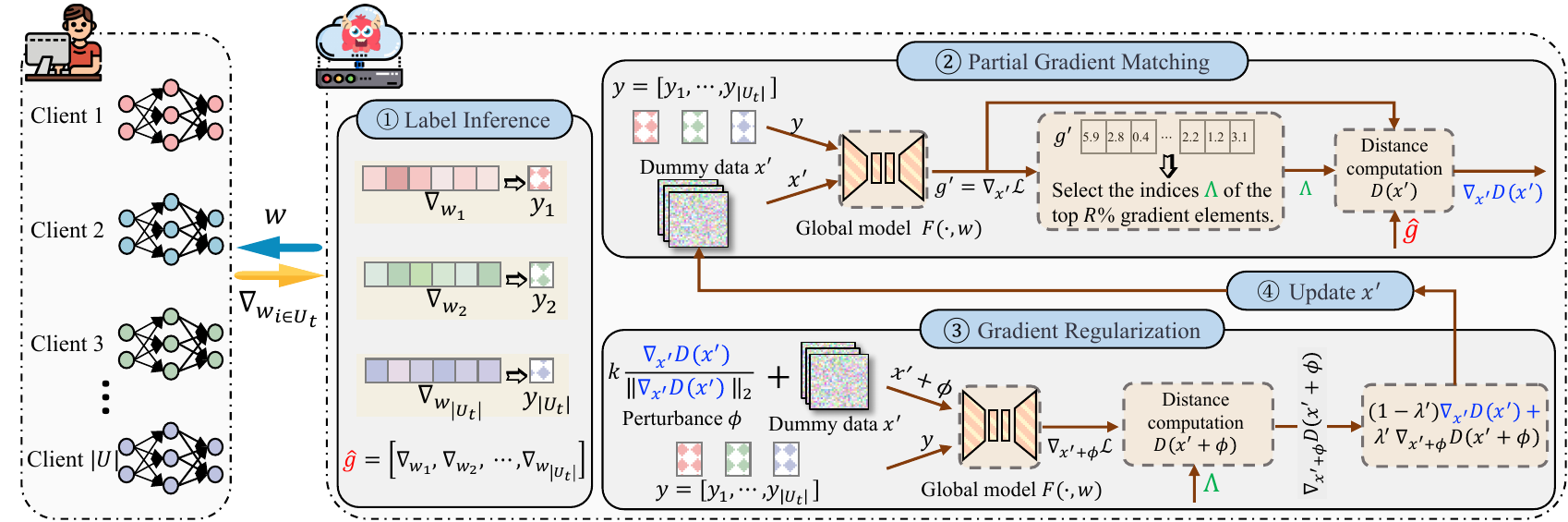}
    \caption{The overview of the proposed attack \sysname.}   
    \label{overview}
\end{figure*}

\subsection{Overview}

To efficiently address the gradient matching problem, \sysname harnesses two novel techniques, partial gradient matching and gradient regularization.
The partial gradient matching selectively matches only specific portions of gradients and can be regarded as a generalized form of gradient matching.
This technique not only significantly reduces $\mu$, but also ensures that the optimal solutions remain consistent with those derived from complete gradient matching.
Moreover, to increase the value of $L$, \sysname incorporates a gradient regularization term into partial gradient matching.
The inclusion of the gradient regularization term, however, requires the computation of Hessian matrix, a computationally intensive task.
To alleviate this burden, we propose an approximation method to enhance the practicality and scalability of \sysname.
Figure~\ref{fig_mu_l_sub2} demonstrates that the combination of the two techniques yields higher and more stable values of $\frac{2\mu}{L}$.
Algorithm~\ref{alg_our} summarizes \sysname and Figure~\ref{overview} presents the entire attack process of \sysname.

\begin{algorithm}[!t]
  \caption{Our attack: \sysname}
  \label{alg_our}
  \begin{algorithmic}[1]
    \Require
    The global model $F$ and parameters $w$;
    the gradients estimated by the server $\hat{g}$;
    the loss function (cross-entropy loss) $\mathcal{L}$;
    the step size $\eta$;
    the blend factor $\lambda'$;
    the matching ratio $R$;
    maximum attack iterations $\mathcal{I}$;
    
    \State Infer the ground-truth labels $y$ from $\hat{g}$ using existing label inference techniques.

    \State Initialize dummy data $x'$ with a uniform distribution ranging from 0 to 1 and $x_0'=x'$.

    \For{$j\gets 0$ to $\mathcal{I}-1$}
        \State Compute the gradients of the loss with respect to dummy data $g_j'= \nabla_{x_j'} \mathcal{L}(F(x_j',w),y)$.
        \State Select the indices of the top $R$\% gradient elements with the largest magnitude in $g_j'$ to form $\Lambda$.
        \State Compute the distance $D(x_j')[\Lambda]$ based on Equation~(\ref{eq_16}).
        \State Compute $\phi = k \frac{\nabla_{x_j'} D(x_j')[\Lambda]}{|| \nabla_{x_j'} D(x_j')[\Lambda] ||_2}$ and the distance $D(x_j'+\phi)[\Lambda]$ based on Equation (\ref{eq_16}).
        \State Mix $\nabla_{x_j'} D(x_j')[\Lambda]$ and $\nabla_{x_j' + \phi} D(x_j' + \phi)[\Lambda]$ using the blend factor $\lambda'$ to update $x_j'$ with step size $\eta$.
        \State Project $x_j'$ onto feasible domain $[0,1]$ to obtain $x_{j+1}'$.
    \EndFor
    
    \State \textbf{Return} the recovered data $x'=x_{\mathcal{I}}'$.
  \end{algorithmic}
\end{algorithm}

\subsection{Less is Better: Partial Gradient Matching}
\label{sec_less_is_better}

The core idea of partial gradient matching is to align only a subset of gradient elements rather than all. 
Consider an index set $\Lambda$ containing the indices of the gradient elements to be matched.
The partial gradient matching problem can be formulated as follows:
\begin{equation}
\label{eq_7}
    x' = \arg \min_{x'} \text{dist}(g'[\Lambda], \hat{g}[\Lambda]).
\end{equation}
When $\Lambda$ contains all gradient elements' indices, the partial gradient matching degenerates into the original gradient matching problem (Equation (\ref{eq_2})).
As illustrated in Theorem~\ref{thm2}, the partial gradient matching problem (Equation (\ref{eq_7})) possesses several favorable properties compared to Equation (\ref{eq_2}).

\begin{thm}
\label{thm2}
Should Equation (\ref{eq_3}) be overdetermined and free of redundant constraints\footnote{Redundant constraints refer to equivalent constraints. This assumption primarily ensures that removing some constraints does not render Equation (\ref{eq_3}) underdetermined. In practice, it is uncommon for the removal of some constraints to significantly alter the solution, allowing us to relax our concerns about this assumption.}, there exists a certain index set $\Lambda$ such that Equation (\ref{eq_7}) not only preserves the same solution as Equation (\ref{eq_2}), but also enjoys a higher value of $\mu$ than that of Equation (\ref{eq_2}), with $\text{dist}(\cdot,\cdot)$ being specified as the $L_1$ distance\footnote{$L_1$ distance evaluates each gradient element uniformly and can facilitate our discussion. Other distance functions are essentially equivalent to considering the gradients with different weights. For example, the $L_2$ distance prioritizes gradient elements that have yet to be well-aligned.}.
\end{thm}

The first part of Theorem~\ref{thm2} asserts that Equations (\ref{eq_2}) and (\ref{eq_7}) yield the same solution, which is obvious from the definition of overdetermined systems.
We mainly demonstrate the second part: Equation (\ref{eq_7}) achieves a higher value of $\mu$ than Equation (\ref{eq_2}).
By setting $\text{dist}(\cdot,\cdot)$ as $L_1$ distance, $\langle x' - x, \nabla_{x'} \text{dist}(g', \hat{g}) \rangle$ can be written as follows:
\begin{IEEEeqnarray}{rCL}
\label{eq_8}
&\langle x' - x, \nabla_{x'} \text{dist}(g', \hat{g})\rangle = \langle x' - x, \frac{1}{|g'|} \frac{\partial}{\partial x'} \sum_i |g'[i] - \hat{g}[i]| \ \rangle \nonumber \\
&= \langle x' - x, \frac{1}{|g'|} \sum_i \text{sign}(g'[i] - \hat{g}[i]) \frac{\partial g'[i]}{\partial x'} \ \rangle.
\end{IEEEeqnarray}
Based on Theorem~\ref{thm2} in Appendix~\ref{appendix_supp_theory}, if $\Lambda$ is chosen to include indices corresponding to the terms with the largest inner product values in Equation (\ref{eq_8}), we arrive at Equation (\ref{eq_9}):
\begin{equation}
\label{eq_9}
\begin{split}
\frac{1}{|\Lambda|} \sum_{i \in \Lambda} \langle x' - x, \text{sign}(g'[i] - \hat{g}[i]) \frac{\partial g'[i]}{\partial x'} \rangle \\
\geq \frac{1}{|\hat{g}|} \sum_i \langle x' - x, \text{sign}(g'[i] - \hat{g}[i]) \frac{\partial g'[i]}{\partial x'} \rangle.
\end{split}
\end{equation}
Equation (\ref{eq_9}) suggests $\langle x' - x, \nabla_{x'} \text{dist}(g'[\Lambda], \hat{g}[\Lambda]) \rangle \geq \langle x' - x, \nabla_{x'} \text{dist}(g', \hat{g}) \rangle$.
Thus, there exists a certain index set $\Lambda$ that yields a higher value $\mu$.

The remaining challenge is how to determine the appropriate $\Lambda$.
While the preceding analysis suggests that $\Lambda$ ought to include indices associated with the largest inner product values, identifying these requires evaluating the gradient for each individual gradient element $g'[i]$ with respect to $x'$, i.e., $\frac{\partial g'[i]}{\partial x'}$.
Given the vast number of parameters in DNNs, this is computationally prohibitive.
To address this issue, we present an approximate solution here.

Generally, within a sufficiently small neighborhood, aligning each gradient element should produce a gradient direction positively correlated with $x' - x$ \cite{convex_optimization}, meaning that each term is usually positive.
As a result, our task reduces to finding the indices $i$ for which $\frac{\partial g'[i]}{\partial x'}$ has bigger magnitude.
Moreover, the term $\frac{\partial g'[i]}{\partial x'}$ with greater magnitude often corresponds to $g'[i]$ with large magnitude.
To clarify, according to the chain rule, there is $\frac{\partial g'[i]}{\partial x'} = \frac{\partial g'[i]}{\partial \mathcal{L}} \frac{\partial \mathcal{L}}{\partial x'}$.
As can be seen, $\frac{\partial \mathcal{L}}{\partial x'}$ remains identical across gradient elements and we focus on the first term, which measures how the gradient changes in response to variations in the loss function.
We can quantify this sensitivity using Fisher information of $g'[i]$, approximated by the square of $g'[i]$ \cite{fisher_info}.
Empirical validation of this can be found in Appendix \ref{appendix_partial}, where the correlation coefficient between the magnitude of gradient elements and their sensitivity to the input data is around 0.7372, a statistically significant value.
To sum up, we match the top $R$\% gradient elements with the largest magnitude in $g'$.

The philosophy behind partial gradient matching is intuitive.
Gradient elements with larger magnitudes correspond to key parameters vital for capturing significant feature information, while gradient elements with smaller magnitudes are related to less influential parameters that primarily contribute noise to the optimization landscape~\citep{backdoor_defense}.

\subsection{Gradient Regularization}

For simplicity, let us define $D(x')$ as $\text{dist}(g'[\Lambda], \hat{g}[\Lambda])$ where $g' = \nabla_{w}\mathcal{L}(F(x', w), y)$.
From Equation (\ref{eq_5}), it is clear that a smaller $\nabla_{x'} D(x')$ along the entire optimization path can enable a smaller value of $L$.
To leverage this insight, we introduce a gradient regularization term in Equation (\ref{eq_7}) as follows:
\begin{equation}
\label{eq_10}
    x' = \arg \min_{x'} \ D(x') + \lambda ||\nabla_{x'}  D(x')||_2,
\end{equation}
where $\lambda$ is a weighting factor to balance the two terms in Equation~\ref{eq_10}.
Gradient-based optimization methods are commonly employed to solve Equation (\ref{eq_10}).
However, as indicated in Equation (\ref{eq_11}), the gradients of Equation (\ref{eq_10}) involve the Hessian matrix evaluated at $x'$, which is computationally prohibitive in high-dimension spaces.
\begin{equation}
\nonumber
\label{eq_11}
    \nabla_{x'} \left( D(x') + \lambda ||\nabla_{x'}  D(x')||_2 \right)  \\
    = \nabla_{x'} D(x') + \lambda  H  \frac{\nabla_{x'}  D(x')}{|| \nabla_{x'}  D(x') ||_2}.
\end{equation}
To circumvent the need for direct Hessian evaluation, we present an approximate estimation for $H  \frac{\nabla_{x'}  D(x')}{|| \nabla_{x'}  D(x') ||_2}$.
Consider a random small perturbance $\phi$ applied to $x'$.
A Taylor expansion of $D(x'+\phi)$ yields:
\begin{equation}
\label{eq_12}
    D(x'+\phi)= D(x')+ \nabla_{x'} D(x')^{\top} \phi.
\end{equation}
Differentiating both sides of Equation (\ref{eq_12}) results in:
\begin{equation}
\label{eq_13}
   \nabla_{x'+\phi} D(x'+\phi)= \nabla_{x'} D(x') + H \phi. 
\end{equation}
Setting $\phi$ in the direction of $\nabla_{x'}  D(x')$ leads to:
\begin{equation}
\label{eq_14}
     H  \frac{\nabla_{x'}  D(x')}{|| \nabla_{x'}  D(x') ||_2} = \frac{\nabla_{x'+\phi} D(x'+\phi) - \nabla_{x'} D(x')}{k}.
\end{equation}
where $\phi=k \frac{\nabla_{x'}  D(x')}{|| \nabla_{x'}  D(x') ||_2}$ and $k$ is a small constant ensuring $\phi$ remains small.
We then substitute Equation~(\ref{eq_14}) into Equation~(\ref{eq_11}):
\begin{equation}
\label{eq_15}
\begin{split}
    &\nabla_{x'} D(x') + \frac{\lambda}{k} (\nabla_{x'+\phi} D(x'+\phi) - \nabla_{x'} D(x')) \\
    &= (1-\frac{\lambda}{k}) \nabla_{x'} D(x') + \frac{\lambda}{k} \nabla_{x'+\phi} D(x'+\phi).
\end{split}
\end{equation}
Equation (\ref{eq_15}) only involves first-order gradients (linear computation time) and is more efficient than direct Hessian evaluation, which demands cubic computation time.
Moreover, if we treat $\frac{\lambda}{k}$ as a constant between 0 and 1, Equation (\ref{eq_15}) essentially blends the gradients generated by $x'$ and $x'+\phi$.
Therefore, we instead set $\lambda'=\frac{\lambda}{k}$ and tune $\lambda'$ within the interval $[0,1]$ to modulate the strength of the gradient regularization.

A closer examination of Equation (\ref{eq_15}) reveals its actual effect in the optimization process.
Notice that, without gradient regularization, each update can be viewed as moving $x'$ a certain distance in the direction of $\phi$. 
When the gradients at $x'$ and $x'+\phi$ oppose each other, it means that the step taken may be excessively large, potentially overshooting the optimal point.
By blending the gradients from both locations, we can effectively dampen the gradient at $x'$, which acts as a form of adaptive step size adjustment.
Thus, it is intuitive that Equation (\ref{eq_15}) indeed improves convergence by dynamically adjusting step sizes based on gradient alignment, allowing for more effective navigation through the complex landscape of the loss function.
We provide an illustrative example in Appendix~\ref{appendix_gr_example}.

\subsection{Concrete Formulation}

\textbf{Loss function.}
Our distance metric consists of four terms, including $L_1$ distance, cosine distance, total variation (TV), and activation value penalty.
Drawing inspiration from~\citep{exploiting_pretained}, we employ both $L_1$ distance and cosine distance to quantify the gradient distance.
The choice of $L_1$ distance is due to its resilience against gradient perturbations, which can occur from various defense methods, making it preferable over $L_2$ distance.
Moreover, TV helps capture differences between neighboring pixels, thereby improving the fidelity of the reconstructed images, given that natural images exhibit smooth transitions in adjacent pixels.
The activation value penalty applies an $L_1$ penalty to the outputs of the model's intermediate layers, promoting sparsity in these activations.
This is based on the observation that pre-trained and subsequently fine-tuned models often produce sparse activations, with many neurons having near-zero values for natural images~\cite{training_survey}.
The ultimate $\text{dist}(\cdot,\cdot)$ is formulated as follows:
\begin{IEEEeqnarray}{rCL}
\label{eq_16}
&\text{dist}(g'[\Lambda], \hat{g}[\Lambda]) = \underbrace{||g'[\Lambda] - \hat{g}[\Lambda]||_1}_{\text{L}_1\text{distance}} + \underbrace{1 - \frac{g'[\Lambda]^{\top} \hat{g}[\Lambda]}{||g'[\Lambda]||_2 ||\hat{g}[\Lambda]||_2}}_{\text{cosine distance}} \nonumber \\
&+ \alpha \cdot \underbrace{\text{TV}(x')}_{\text{total variation}} + \beta \cdot \sum_{l} \underbrace{ ||F^{(l)}(x', w)||_1}_{\text{activation value penalty}}, 
\end{IEEEeqnarray}
where $F^{(l)}(x', w)$ denotes the activation values of the $l$-th layer of the model given the input $x'$.
$\alpha$ and $\beta$ are used to balance the influence of the two terms.

\textbf{Optimization.}
Any gradient-based optimization method can be employed to address Equation (\ref{eq_16}).
However, there is an additional constraint: pixel values are required to be limited within the range $[0,1]$.
To adhere to this constraint, we include a projection step, as shown in the 9th step of Algorithm~\ref{alg_our}, clamping pixel values to this range.

\section{Evaluation Protocol}
\label{sec_setup}
\renewcommand{\arraystretch}{1.15}
\begin{table*}[!ht]
\centering
\caption{The configurations considered by existing attack methods. When multiple settings are employed, we report the most sophisticated configuration. The ConvNet is a small CNN developed by~\citet{dlg}. \citet{gradinversion} utilized a batch size of 48, but the resulting reconstructions are of low fidelity. CNNs indicate the use of various CNN architectures. Some methods employ ImageNet but resize images from 224 to either 64 or 128.  "Early" and "Late" indicate when the attack occurs: during the first communication round and late training phase, respectively. "Default" refers to initialization methods provided by PyTorch while "Wide Uniform" indicates the use of a wide uniform distribution for initializing model parameters. Instances where authors do not provide specific details are marked as "N/A," and a "$-$" indicates no auxiliary information used.}
\label{eval_setup_comp}
\scriptsize
\begin{tabular}{@{}cccccccc@{}}
\toprule
Evaluation Setting &
  \begin{tabular}[c]{@{}c@{}}Batch\\ Size\end{tabular} &
  Model &
  Dataset &
  Attack Stage &
  \begin{tabular}[c]{@{}c@{}}Initialization\\ Method\end{tabular} &Auxiliary Information & \begin{tabular}[c]{@{}c@{}}Local\\ Step\end{tabular}\\ \midrule
\citet{dlg} & $\leq$ 8  & ConvNet & Small Datasets & Early & Wide Uniform & $-$ & 1 \\
\citet{idlg} & $\leq$ 8   & ConvNet & Small Datasets & Early & Wide Uniform & $-$ & 1  \\
\citet{invertinggrad} & 1 & ResNet & ImageNet & Early & Pretrained & $-$ & 1 \\
\citet{gradinversion} & $\leq$ 48  & ResNet & ImageNet & Early & Pretrained & Batch Normalization Statistics & 1  \\
\citet{bayes_attack} & 1 & ConvNet & Small Datasets & Early \& Late & Wide Uniform & $-$ & 1 \\
\citet{grad_gene1} & $\leq$ 48 & ResNet & ImageNet ($64\times64$) & Early & Default & Clients' Data Distribution & 1 \\
\citet{grad_gene2} & 1 & ResNet & ImageNet & Early & Default & Clients' Data Distribution & 1 \\
\citet{grad_gene3} & 16 & CNNs & ImageNet ($128\times128$) & Early & N/A & Clients' Data Distribution & 5 \\
\citet{LGLA} & 16 & CNNs &  Small Datasets & Early & N/A & $-$ & 5 \\ \hline
Ours & $\leq$ 64 & CNNs \& Transformers & ImageNet \& Medical Datasets & Early \& Late & Default \& Pretrained & $-$ & $12$  \\ \bottomrule
\end{tabular}
\end{table*}

Current evaluation protocols do not adequately capture the nuances of real-world FL environments.
To fill this gap, we develop an evaluation protocol that faithfully reflects production environments, drawing from our investigation (Appendix~\ref{appendix_investigation}).
We identify eight critical factors that significantly influence attack performance and align their values with those found in production environments, establishing a more practical evaluation criterion.
Table~\ref{eval_setup_comp} provides a comparison of their values in previous studies versus those in this paper:
\begin{itemize}[leftmargin=*,topsep=0pt]

    \item \textbf{Batch size.} The batch size determines the number of optimization variables included in the gradient matching problem. When the batch size increases, there are more variables to optimize, making it harder to reconstruct clients' data. Appendix~\ref{appendix_investigation} shows that practical batch sizes typically range from 16 to 64.

    \item \textbf{Model.} Beyond CNNs, transformer models are emerging as competitive alternatives. Appendix~\ref{appendix_investigation} illustrates the usage frequency of both types in practical applications, revealing that CNNs are predominant while transformer models are less common. Thus, although transformer models will be evaluated, the primary focus remains on CNNs. We evaluate four distinct models: ResNet, DenseNet, MobileNet, and Vision Transformer (ViT). The first three are based on convolutional structures, and ViT is based on transformers. Moreover, to provide a comprehensive evaluation, we employ variants of ResNet with 10, 18, and 34 layers, as well as expanded channel widths for ResNet10, labeled as ResNet10\_x3 and ResNet10\_x5, to study how model complexity impacts attack performance.

    \item \textbf{Dataset selection.} Real-world datasets are more sophisticated than commonly used small benchmark datasets, such as MNIST, SVHN, CIFAR-10, and CIFAR-100. Alongside benchmark datasets, we include three more sophisticated datasets, ImageNet, HAM10000, and Lung-Colon Cancer. The latter two are noteworthy as they pertain to medical imaging. HAM10000 contains 10000 dermatoscopic images of pigmented skin lesions across seven classes, while Lung-Colon Cancer features 25000 histopathological images categorized into five classes. Given that FL is primarily applied in privacy-sensitive fields like healthcare, and considering the substantial differences between medical images and common images, evaluating GLAs in HAM10000 and Lung-Colon Cancer can provide a clearer picture of their practical implications.

    \item \textbf{Attack stage.} Attacks can occur at any communication round in FL. However, the quality of reconstructed data is significantly higher in the initial communication rounds compared to later ones~\cite{bayes_attack,sok_grad_leakage}. Note that clients of interest may not participate in the initial communication rounds. Therefore, rather than focusing solely on the attack performance of GLAs in the initial rounds, we evaluate their performance both at the onset and during later phases of FL to better understand the overall privacy leakage risk associated with FL.

    \item \textbf{Initialization method.} Some studies~\cite{bayes_attack} found that GLAs perform better in untrained models with parameters initialized through wide uniform distributions. However, Appendix~\ref{appendix_investigation} illustrates that, in practice, about 70\% of cases use default random initialization, while a smaller percentage employs pre-trained weights. We focus on default random initialization and pre-trained weights. 

    \item\textbf{Auxiliary information.} We argue against the use of auxiliary information for two reasons. First, such information may not always be available in real-world scenarios~\cite{eval_frame1,out}, limiting the generalizability of attack methods depending on it. Second, methods with auxiliary information often incur high attack costs and complexities, making them less practical to implement. At the attack design level, \sysname does not utilize such information.

    \item \textbf{Local step.} To reduce communication costs, clients can perform several local training steps. More local steps mean that model updates sent to the server are based on a larger amount of data. Also, the gradients estimated by the server are cumulative from each local step, rather than reflecting the gradients from all relevant data points, leading to estimation errors. Both of these two aspects make it harder to recover clients' data. Appendix~\ref{appendix_investigation} indicates that in real-world environments, using a local step of 1 and training for one epoch locally are the two most common configurations.

    \item \textbf{Defenses.} In real-world scenarios, as shown in Appendix~\ref{appendix_investigation}, most studies avoid cryptography-based methods due to their substantial computational costs. Instead, the majority of studies favor Gaussian-DP~\citep{dp}. In addition to DP, we also include four state-of-the-art defense methods: quantization~\citep{grad_gene3}, Soteria~\citep{soteria2}, OUTPOST~\citep{out}, and Guardian~\citep{guardian}.

\end{itemize}

\begin{table*}[!t]
\caption{A summary of the configurations across different sections along with corresponding privacy leakage risks. In particular, Section 6.3 includes four configuration changes, labeled \ding{172} through \ding{175}, with red indicating weaker configurations to be discarded. \ding{173} and \ding{175} involve multiple variables. Moreover, in \ding{175}, clients have the option to perform a local step of 1 with batch sizes of 16, 32, or 64, or alternatively, they can train for one local epoch using a batch size of 16 (which corresponds to 12 local steps in the Lung-Colon Cancer dataset). The latter is more challenging. The risk levels are categorized into three levels: Very High (PSNR of $\geq 20$), High (PSNR of $15 \sim 20$), and Medium (PSNR of $10 \sim 15$), which indicate that nearly all, most, and only a small portion of clients' data can be semantically reconstructed, respectively.}
\label{quick_check}
\scriptsize
\centering
\begin{tabular}{@{}cccccccccc@{}}
\toprule
Config. & Batch & Model    & Dataset                               & Attack Stage & Initialization & Heterogeneity & Local Step & Defense & Risk \\ \midrule
Sec. \ref{eval_benchmark}      & 16,32,64   & ResNet10 & Benchmark Datasets         & Early \& Late & Pretrained            & IID                & 1          & \ding{55}     & Very High   \\
Sec. \ref{eval_complex_dataset}      & 16,32,64   & ResNet10 & Sophisticated Datasets & Early \& Late & Pretrained            & IID                & 1          & \ding{55}     & High to Very High   \\
Sec. \ref{eval_diverse_settings} & \begin{tabular}[c]{@{}c@{}}(\textcolor{red}{\ding{175} 16,32,64})\\ vs. 16\end{tabular} & \begin{tabular}[c]{@{}c@{}}(\textcolor{red}{\ding{174} Others})\\ vs. ResNet10\end{tabular} & Lung-Colon Cancer             & \begin{tabular}[c]{@{}c@{}}(\textcolor{red}{\ding{173} Early})\\ vs. Early \& Late\end{tabular} & \begin{tabular}[c]{@{}c@{}}(\textcolor{red}{\ding{173} Random})\\ vs. Pretrained\end{tabular} & \begin{tabular}[c]{@{}c@{}}(\textcolor{red}{\ding{172} Non-IID})\\ vs. IID\end{tabular} & \begin{tabular}[c]{@{}c@{}} (\textcolor{red}{\ding{175} 1}) \\ vs. 12 \end{tabular} & \ding{55}     & Medium to High   \\
Sec. \ref{eval_defense}      & 16         & ResNet10 & Lung-Colon Cancer                     & Early \& Late & Pretrained            & IID                & 12       & \ding{52}     & Medium   \\ \bottomrule
\end{tabular}
\end{table*}

\begin{table*}[!ht]
\centering
\caption{The PSNRs of reconstructed images using four attack methods against FL configurations with benchmark datasets and varying batch sizes. The best results are given in bold.}
\label{table_main}
\scriptsize
\begin{tabular}{@{}c|cccc|cccc|cccc|cccc@{}}
\toprule
Dataset    & \multicolumn{4}{c|}{MNIST}    & \multicolumn{4}{c|}{SVHN}     & \multicolumn{4}{c|}{CIFAR-10} & \multicolumn{4}{c}{CIFAR-100} \\ \midrule
Batch Size & iDLG  & IG    & GI    & Ours   & iDLG  & IG    & GI    & Ours   & iDLG  & IG    & GI    & Ours   & iDLG  & IG    & GI    & Ours   \\ \midrule
16         & 10.70 & 10.96 & 11.57 & \textbf{22.52} & 10.04 & 10.28 & 11.43 & \textbf{21.62} & 9.90  & 10.26 & 11.26 & \textbf{23.06} & 9.86  & 10.00 & 11.42 & \textbf{22.94} \\
32         & 9.39  & 9.96  & 11.17 & \textbf{21.57} & 9.53  & 9.43  & 10.41 & \textbf{20.11} & 9.51  & 9.58  & 10.59 & \textbf{21.33} & 9.39  & 9.88  & 10.48 & \textbf{21.50} \\
64         & 9.27  & 9.88  & 10.94 & \textbf{21.12} & 8.86  & 9.37  & 10.17 & \textbf{18.09} & 8.52  & 9.42  & 10.01 & \textbf{19.83} & 8.58  & 9.40  & 10.26 & \textbf{19.38} \\ \bottomrule
\end{tabular}
\end{table*}

\textbf{Evaluation approach.}
Our evaluation begins with a seed configuration using ResNet10, small benchmark datasets, ImageNet-pretrained weights, and a local step of 1.
We then progressively explore more intricate configurations by changing one factor at a time and keeping track of the most challenging values encountered.

\textbf{Generic setup for FL.} The remaining FL configurations largely follow existing works~\cite{soteria,soteria2,guardian}.
We utilize an FL scenario with 100 clients collaborating to train a global model.
We consider both IID and Non-IID settings.
In the IID setting, the training dataset is randomly partitioned into 100 equal subsets, one per client.
See Section \ref{eval_diverse_settings} for Non-IID settings.
The server performs average aggregation, and each client updates its local model with a learning rate of $1 \times 10^{-4}$.

\textbf{Attacks.}
We select three state-of-the-art attacks, iDLG~\citep{idlg}, InvertingGrad (IG)~\citep{invertinggrad}, and GradInversion (GI)~\citep{gradinversion} as baselines.
Additionally, although we assume that clients' data distributions are inaccessible, we also include comparisons with generative model-based methods, i.e., GGL~\cite{grad_gene2} and ROGS~\cite{grad_gene3}, to demonstrate the effectiveness of \sysname.

\textbf{Evaluation metrics.}
To assess the similarity between reconstructed images and their original counterparts, four metrics are used, including PSNR ($\uparrow$), SSIM ($\uparrow$), LPIPS ($\downarrow$), and attack success rate (ASR) ($\uparrow$).
Here, ($\uparrow$) indicates that larger values represent better results and ($\downarrow$) signifies the opposite.
PSNR measures logarithmic $L_2$ distance between reconstructed and original images.
SSIM quantifies the structural similarity between two images.
LPIPS measures perceptual similarity by comparing DNN-extracted features.
For ASR, we enlist three human volunteers to determine whether the entities in the two images have a high semantic overlap and report the proportion of instances with strong overlap.

\begin{figure*}[!t]
    \centering
    \subfloat[ImageNet]{
        \includegraphics[width=0.3\linewidth]{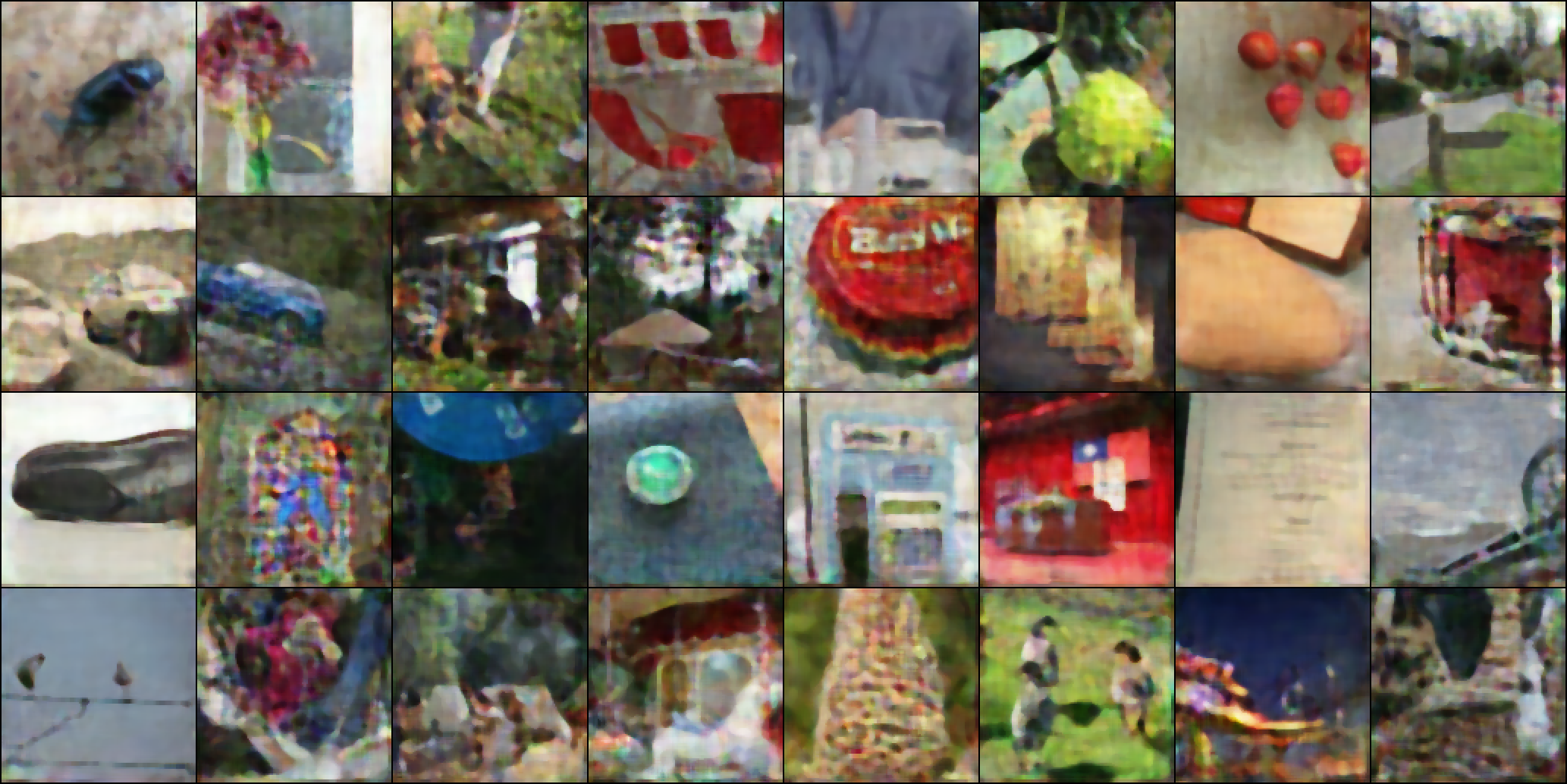}
    }
    \subfloat[HAM10000]{
        \includegraphics[width=0.3\linewidth]{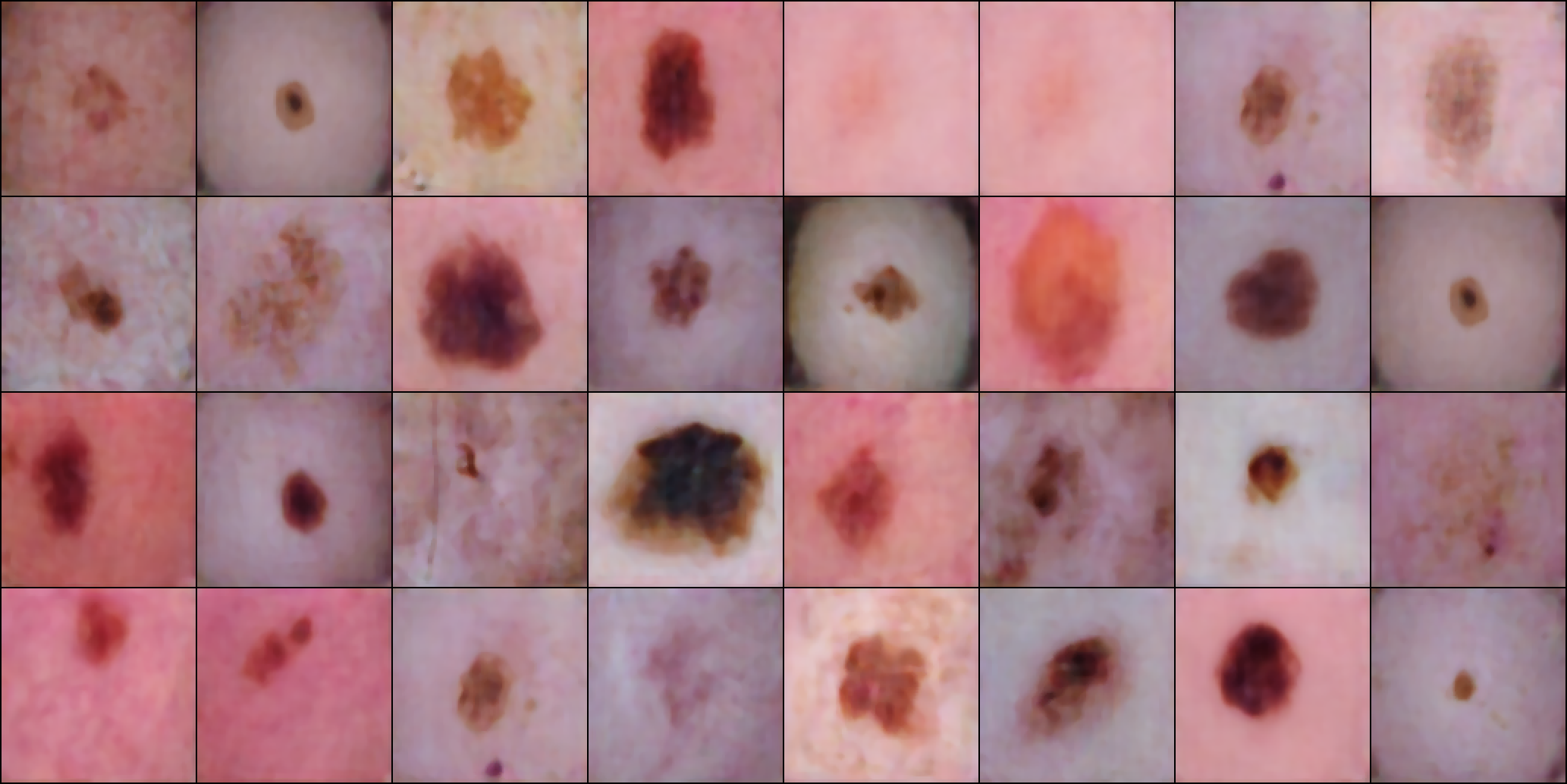}
    }
    \subfloat[Lung-Colon Cancer]{
        \includegraphics[width=0.3\linewidth]{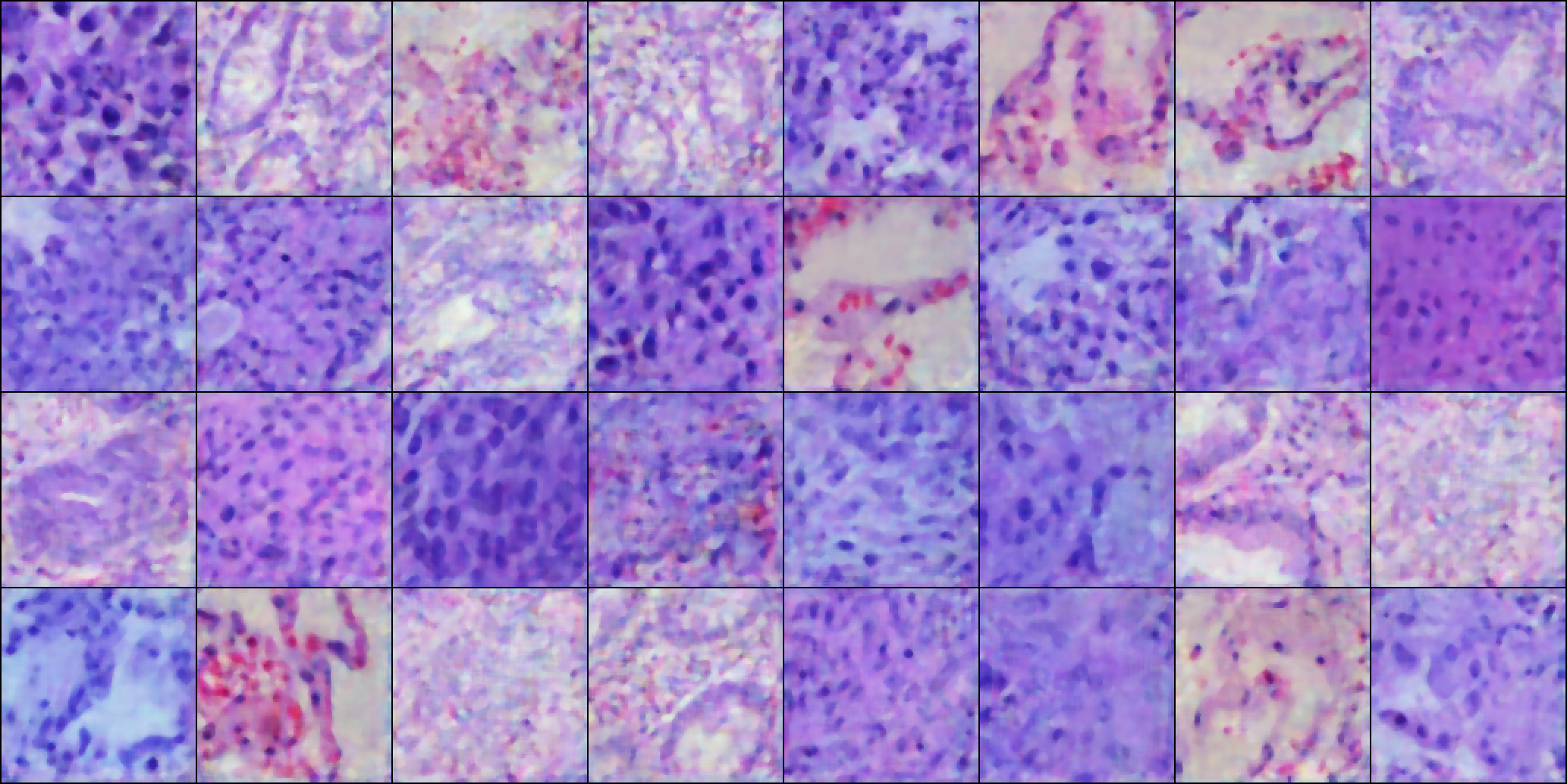}
    }
    \caption{The images recovered by \sysname in three datasets. The batch size here is 32. See Figure \ref{appendix_real_imgs} for ground-truth ones and Figure \ref{ROGS_recovered} for reconstructions of ROGS.
    }
    \label{visual_complex_dataset}
\end{figure*}

\revise{\textbf{Label inference.}
We infer ground-truth labels by aggregating gradients from the final fully connected layer along the feature dimension and identifying the indices corresponding to minimum values~\citep{gradinversion}.
Formally, let $\Delta W \in \mathbb{R}^{K \times N}$ denote the gradient matrix where $K$ is the feature dimension and $N$ is the total number of classes.
The inferred labels are computed as: $y' = \text{arg min}( \sum_{i=1}^{K} \Delta W[i,:])) [:\text{batch size}]$.
If the batch size exceeds the number of classes, we infer labels based on gradient magnitudes.
The count for each class $j$ is calculated as:
$c_j = \lfloor \text{batch size} \times \frac{\sum_{i=1}^{K} (\Delta W[i,j] - \max{W})}{\sum_{i=1}^{K} \sum_{j=1}^{N} (\Delta W[i,j] - \max{W})} \rfloor$.
If $\sum c_j < \text{batch size}$, we supplement the batch by iteratively adding samples from the class with the lowest gradient strength until the batch size requirement is satisfied.
We also jointly optimize the inferred labels alongside the input images during the inversion process.
Appendix \ref{appendix_sens_ablation} examines the impact of label inference.}

\textbf{Hyperparameters and others.}
Adam optimizer is utilized for Equation (\ref{eq_16}) with step size $\eta=1 \times 10^{-4}$ and attack iterations $\mathcal{I}=10000$.
The hyperparameters $\alpha$, $\beta$, and $\lambda'$ for \sysname are configured to $1 \times 10^{-5}$, $1 \times 10^{-4}$, and $0.7$, respectively.
We match the most significant 50\% of gradient elements, i.e., $R=50$.
Default hyperparameters are used for alternative attack methods, with attack iterations increased to 10000 for convergence.
Attacks are carried out at the 2000$\times \{0,1,2,3,4,5\}$-th rounds.
For each image, we find the most similar reconstructed image to calculate metrics.

\section{Evaluation Results}
\label{sec_exp}

\begin{table*}[!t]
\caption{The PSNRs of reconstructed images using three attack methods against FL configurations with sophisticated datasets. }
\label{table_main2}
\centering
\scriptsize
\begin{tabular}{@{}c|ccc|ccc|ccc@{}}
\toprule
Dataset & \multicolumn{3}{c|}{ImageNet} & \multicolumn{3}{c|}{HAM10000} & \multicolumn{3}{c}{Lung-Colon Cancer} \\ \midrule
Batch Size & GGL   & ROGS  & Ours           & GGL   & ROGS  & Ours           & GGL   & ROGS  & Ours           \\ \midrule
16         & 12.50 & 16.26 & \textbf{20.07} & 11.76 & 14.54 & \textbf{23.35} & 10.81 & 11.33 & \textbf{18.40} \\
32         & 11.31 & 14.18 & \textbf{19.07} & 10.61 & 13.28 & \textbf{22.47} & 9.86  & 10.44 & \textbf{17.73} \\
64         & 10.17 & 12.96 & \textbf{19.01} & 9.81  & 11.47 & \textbf{22.16} & 9.48  & 10.27 & \textbf{16.06} \\ \bottomrule
\end{tabular}
\end{table*}

\begin{table}[!th]
\caption{\revise{The performance of \sysname under different non-IID scenarios in Lung-Colon Cancer. We use a batch size of 32. For feature skew, $0 \sim 20$\% is the most simple group.}}
\label{table_skew}
\centering
\scriptsize
\begin{tabular}{@{}c|ccccc@{}}
\toprule
Label Skew    & 1 class     & 2 class & 3 class & 4 class  & 5 class \\
PSNR          & 18.82          & 18.38    & 17.97    & 17.85    & 17.73     \\ \midrule
Quantity Skew & 50          & 100   & 150   & 200   & 250    \\
PSNR          & 17.80          & 17.71    & 17.76    & 17.72    & 17.73     \\ \midrule
Feature Skew  & 0-20\% & 20-40\% & 40-60\% & 60-80\% & 80-100\% \\
PSNR          & 19.64          & 18.56    & 17.84    & 16.94    & 16.07     \\ \bottomrule
\end{tabular}
\end{table}

\begin{table}[!th]
\caption{The performance of \sysname under varying data heterogeneity levels.}
\label{table_data_heter}
\centering
\scriptsize
\begin{tabular}{@{}ccccc@{}}
\toprule
$\alpha$ & 0.1   & 0.5   & 1     & 5     \\ \midrule
ROGS  & 10.93 & 10.78 & 10.40 & 10.38 \\
Ours  & 18.06 & 17.86 & 17.65 & 17.60 \\ \bottomrule
\end{tabular}
\end{table}

\begin{table}[!th]
\centering
\caption{The PSNRs of reconstructed images with different weight initialization methods in Lung-Colon Cancer.}
\label{table_weight}
\scriptsize
\begin{tabular}{@{}c|cc|cc@{}}
\toprule
Batch Size & \multicolumn{2}{c|}{32}         & \multicolumn{2}{c}{64}          \\ \midrule
Weight     & Random         & Pre-trained     & Random         & Pre-trained     \\ \midrule
ROGS       & 13.95          & 10.44          & 12.04          & 10.27          \\
Ours       & \textbf{19.62} & \textbf{17.73} & \textbf{18.32} & \textbf{16.06} \\ \bottomrule
\end{tabular}
\end{table}

\subsection{Attack in Benchmark Datasets}
\label{eval_benchmark}

Table~\ref{table_main} presents the PSNR of reconstructed images using four different attack methods, evaluated across various datasets and batch sizes.
Throughout all datasets and batch sizes, \sysname consistently outperforms existing attack methods.
For instance, in CIFAR-10 with a batch size of 16, \sysname achieves a PSNR of 23.06, while the other methods lag far behind with PSNRs of 9.90, 10.26, and 11.26, respectively.

\subsection{Attack in Sophisticated Datasets}
\label{eval_complex_dataset}

We now evaluate the performance of the attack methods on sophisticated datasets, including ImageNet, HAM10000, and Lung-Colon Cancer.
Existing attack methods, including iDLG, IG, and GI, generally exhibit poor reconstruction quality on high-resolution images.
To facilitate a quantifiable comparison, we contrast \sysname with GGL and ROGS.
For HAM10000 and Lung-Colon Cancer, we resize input images to $224 \times 224$.
Table~\ref{table_main2} reports the performance of the three methods across these sophisticated datasets.

Overall, \sysname significantly outperforms GGL and ROGS by a considerable margin.
For example, in HAM10000, \sysname achieves a PSNR of at least 22.16, while GGL and ROGS reach a maximum PSNR of only 14.54.
Crucially, the impressive attack performance of \sysname is attained without needing access to clients' data distribution, whereas both GGL and ROGS do require such access.
Figure~\ref{visual_complex_dataset} shows reconstructed batches for all three datasets.
We also note that the attack methods tend to be less effective on the Lung-Colon Cancer dataset compared to the others. 
This is likely due to the intricate structures, e.g., various tissue patterns and cellular details, presented in Lung-Colon Cancer, as shown in Figure~\ref{visual_complex_dataset}.
In contrast, the images in HAM10000 are relatively simpler and lack such complex structural information, making them easier to reconstruct.
To conduct a more rigorous evaluation of \sysname's attack performance, our subsequent experiments will employ Lung-Colon Cancer.

\begin{table*}[!ht]
\scriptsize
\centering
\caption{The PSNRs of reconstructed images with different models in Lung-Colon Cancer.}
\label{table_model_impact}
\begin{tabular}{@{}cccccccccc@{}}
\toprule
\multirow{2}{*}{Batch Size} & \multirow{2}{*}{Attack} & \multirow{2}{*}{ResNet10} & \multicolumn{2}{c}{Depth}       & \multicolumn{2}{c}{Width}       & \multicolumn{3}{c}{Architecture}                 \\ \cmidrule(l){4-5} \cmidrule(l){6-7} \cmidrule(l){8-10} 
                            &                         &                           & ResNet18       & ResNet34       & ResNet10\_x3   & ResNet10\_x5   & DenseNet121    & MobileNetV2    & ViT            \\ \midrule
\multirow{2}{*}{32}         & ROGS                    & 10.44                     & 10.66          & 11.15          & 10.96          & 11.12          & 10.22          & 9.48           & 9.01           \\
                            & Ours                    & \textbf{17.73}            & \textbf{18.15} & \textbf{18.92} & \textbf{18.46} & \textbf{19.41} & \textbf{17.42} & \textbf{16.85} & \textbf{15.64} \\ \midrule
\multirow{2}{*}{64}         & ROGS                    & 10.27                     & 10.73          & 11.22          & 11.43          & 11.68          & 9.90            & 9.19           & 8.57           \\
                            & Ours                    & \textbf{16.06}            & \textbf{16.94} & \textbf{17.98} & \textbf{16.64} & \textbf{18.18} & \textbf{15.50}  & \textbf{15.20}  & \textbf{13.98} \\ \bottomrule
\end{tabular}
\end{table*}

\begin{table}[!ht]
\centering
\caption{The PSNRs of reconstructed images with varying local steps in Lung-Colon Cancer.}
\label{table_localstep}
\scriptsize
\begin{tabular}{@{}ccccc@{}}
\toprule
Local Step & IG   & GI    & ROGS  & Ours  \\ \midrule
1          & 8.82 & 10.12 & 11.33 & \textbf{18.40} \\
2          & 8.44 & 9.86  & 10.98 & \textbf{16.74} \\
4          & 8.53 & 9.53  & 10.88 & \textbf{15.57} \\
8          & 8.28 & 9.41  & 10.74 & \textbf{14.34} \\
12         & 8.59 & 9.10  & 10.43 & \textbf{13.55} \\ \bottomrule
\end{tabular}
\end{table}

\begin{table}[!ht]
\centering
\caption{The quality of reconstructed images assessed by three distinct evaluation metrics in Lung-Colon Cancer.}
\label{table_metric}
\scriptsize
\begin{tabular}{@{}c|ccc@{}}
\toprule
Metric & SSIM          & LPIPS           & ASR (\%)        \\ \midrule
ROGS   & 0.55          & 0.4736          & 0.35 \%            \\
Ours   & \textbf{0.68} & \textbf{0.3447} & \textbf{20.40\%} \\ \bottomrule
\end{tabular}
\end{table}

\subsection{Attack in Diverse Settings}
\label{eval_diverse_settings}

In this subsection, we examine how the data heterogeneity among clients, the use of pre-trained weights, and the choice of models affect attack performance.
We will subsequently increase the number of local steps to create more challenging attack configurations.
Finally, to mitigate the risk of \sysname overfitting to a single evaluation metric, we incorporate three supplementary metrics for a more comprehensive assessment.

\revise{\textbf{Evaluation in Non-IID settings.}
We use Lung-Colon Cancer and stimulate three non-IID scenarios, including label skew (1-5 classes per client), quantity skew (allocating local datasets of $50 \sim 250$ samples), and feature skew (partitioning images into five percentile groups based on their TV values, with lower TV indicating simpler images).
As shown in Table~\ref{table_skew}, \sysname maintains strong attack performance across all non-IID scenarios.
Notably, higher label skew (e.g., single-class clients) and simpler features (low TV) are more vulnerable to \sysname, while sample quantity variations show negligible impact.
In Table~\ref{table_data_heter}, we further test \sysname with different levels of data heterogeneity by varying Dirichlet distribution $\alpha$ values.
A smaller $\alpha$ indicates stronger data heterogeneity \footnote{\url{https://github.com/TsingZ0/PFLlib}.}.
We see that data heterogeneity does have some impact on the performance of GLAs, but this effect is not particularly significant.
Actually, data heterogeneity mainly affects model performance, with a lesser effect on the gradient matching problem.
Therefore, we will retain the current IID data distribution configuration.}

\textbf{Evaluation in randomly initialized models.}
Table~\ref{table_weight} reports the performance of GLAs in the first training round in models with both randomly initialized and pre-trained weights.
As anticipated, GLAs indeed present higher attack performance against randomly initialized weights.
The reason has been intuitively discussed in Section~\ref{sec_introduction}.
The attack results in later training phases are not reported since we observe negligible differences in attack effectiveness after several training rounds.
Given that pre-trained weights offer greater resistance against GLAs, we will continue to use them moving forward.

\textbf{Evaluation over different model architectures, depths, and widths.}
We evaluate the performance of \sysname from three perspectives: model architecture, depth, and width.
The results are detailed in Table~\ref{table_model_impact}.
Across diverse model architectures, depths, and widths, \sysname consistently surpasses the performance of ROGS.
We note that larger models exacerbate privacy leakage, aligning with conclusions drawn by \citep{invertinggrad,memeory1,memeory2,memeory3}.
Specifically, for \sysname, increasing the model depth from 10 to 34 layers leads to PSNR improvements of 1.19 and 1.92 for batch sizes of 32 and 64, respectively.
This trend is even more pronounced with model width, yielding enhancements of 1.68 and 2.12 in PSNR when transitioning from ResNet10 to ResNet10\_x5.
In terms of architecture, \sysname performs less effectively with MobileNetV2 and ViT compared to ResNet and DenseNet.
The reduced performance with MobileNetV2 is due to its lightweight design, while ViT's unique architecture and information processing pose challenges for \sysname.
Nonetheless, we highlight that \sysname achieves significant attack performance, particularly in comparison to ROGS which typically obtains PSNRs around 10.
Given that CNNs are more prevalent in practical scenarios and less data-hungry, we decide to stick with ResNet10.
We will also see that ResNet10 can attain high accuracy on Lung-Colon Cancer in Section~\ref{eval_defense}, eliminating the need for larger and potentially more complex models.

\textbf{Evaluation over varying local step sizes.}
Here, we examine the case where the local epoch is set to 1.
Lung-Colon Cancer comprises 25000 samples, with 5000 samples reserved for the test set, resulting in each client having a local dataset of 200 samples.
Using a batch size of 16\footnote{This represents a worst-case scenario, as a small batch size for a fixed number of training samples indicates a higher number of local steps, potentially increasing the server's gradient estimation error.
But excessively small batch sizes, e.g., a batch size of 1, may lead to model overfitting on the local dataset.}, each participating client performs $\frac{200}{16} = 12.5$ local steps in a training round.
For convenience, we round the local step size at 12.
Table~\ref{table_localstep} presents the performance of four attack methods when clients execute 1, 2, 4, 8, and, 12 local steps using a batch size of 16.
A key observation is that \sysname achieves better performance over ROGS.
Moreover, as expected, the performance of all attack methods declines as the local step increases.
However, commonly used local steps, either one step or a single local epoch, are insufficient to safeguard against \sysname.
We will use a local step size of 12 for further analysis.

\textbf{Evaluation over diverse metrics.}
Table~\ref{table_metric} reports the performance of ROGS and \sysname, scrutinized through three distinct metrics.
Due to the high cost of manual ASR evaluation, we only calculate ASRs for data reconstructed at the first client during $2000 \times \{0, 1, 2, 3, 4, 5\}$-th training rounds.
The results in Table~\ref{table_metric} demonstrate the superiority of \sysname over ROGS across all three metrics.
Specifically, the SSIM for reconstructed data using \sysname reaches an impressive score of 0.68 (68\% similarity). 
Furthermore, human evaluation reveals an ASR of 20.40\% for \sysname.
These numbers underscore the significant privacy vulnerability in FL.

\begin{figure}[!t]
    \centering
    \subfloat{
        \includegraphics[width=0.9\linewidth]{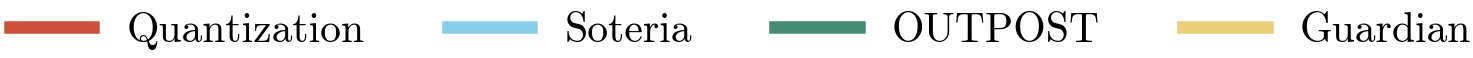}
    } \\
    \vspace{-0.35cm}
    \subfloat[Local Step of 1]{
        \includegraphics[width=0.48\linewidth]{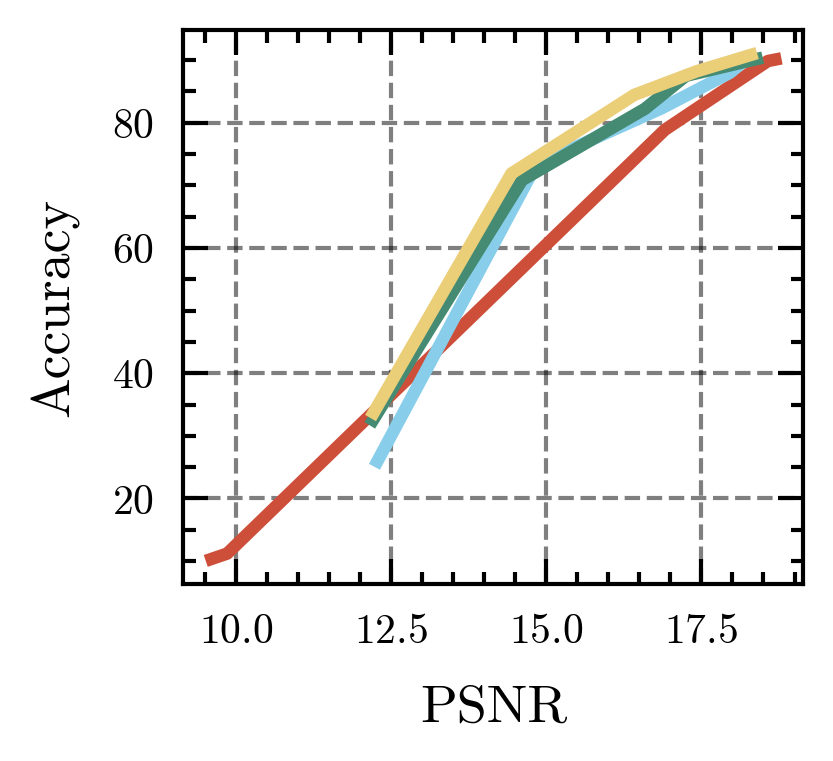}
    }
    \subfloat[Local Step of 12]{
        \includegraphics[width=0.48\linewidth]{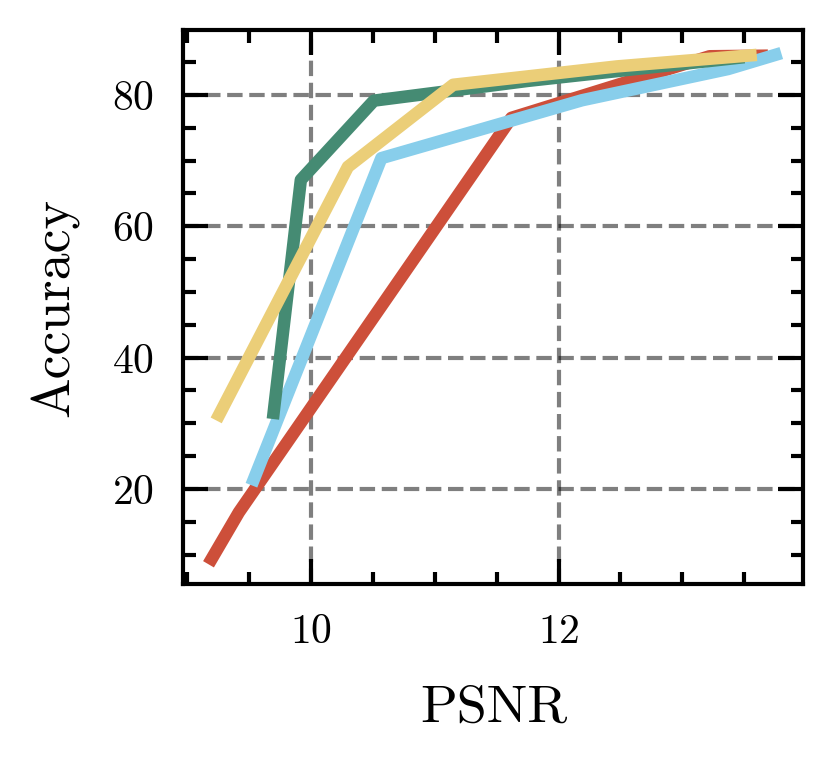}
    }
    \caption{The utility-privacy trade-offs of various defenses in Lung-Colon Cancer against \sysname. }
    \label{fig_trade_off}
\end{figure}

\begin{figure}[!t]
    \centering
    \includegraphics[width=0.85\linewidth]{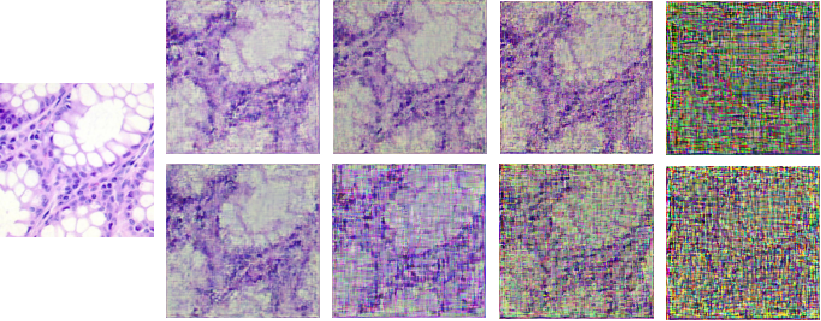}
    \caption{\revise{Recovered images of \sysname against combined defenses of Soteria and DP in Lung-Colon-Cancer. The leftmost image is the original image. The top row represents adaptive DP, while the bottom row shows standard DP-SGD. The values of $\epsilon$ from left to right are $\{ 10^5, 10^4, 10^3, 10^2 \}$.}}
    \label{imagenet_example_defense}
\end{figure}

\subsection{Attack in State-of-the-art Defenses}
\label{eval_defense}

We now employ perturbation-based defenses to construct the most challenging configuration.
Although large defense strengths can render GLAs ineffective, they also cause severe performance degradation. 
In light of this, we adjust defense strengths to plot the utility-privacy trade-off curve to evaluate the performance of GLAs against these defenses.

\textbf{Utility-privacy trade-offs.}
We examine two configurations: local steps of 1 and 12.
Figure~\ref{fig_trade_off} illustrates the utility-privacy trade-offs (accuracy-PSNR curves) of different defenses in Lung-Colon Cancer.
For Soteria and OUTPOST, the pruning ratios used are $\{0.1, 0.3, 0.5, 0.7, 0.9\}$.
Quantization is assessed with bit levels $\{32, 16, 8, 4, 1\}$, and Guardian's effectiveness is evaluated with $\beta$ of $\{0.1, 1, 10, 10^2, 10^3\}$.
Since other attacks already exhibit poor performance without defense, their results are no longer reported.
In Figure~\ref{fig_trade_off}, all defenses require a trade-off in model accuracy to reduce PSNR.
Generally, a PSNR below 10 indicates that the reconstructed data no longer retains meaningful information about the original data~\cite{guardian}.
With a local step of 1, few methods achieve full privacy protection, except for Quantization.
While Quantization lowers PSNR below 10, it severely compromises model performance, resulting in only 10\% accuracy.
Increasing the local step from 1 to 12 can significantly enhance the defense effectiveness, allowing us to maintain a model accuracy of around 50\% while achieving a PSNR of 10.
However, this accuracy remains unsatisfactory.
These findings underscore achieving satisfactory model performance while preventing privacy leakage still remains a significant challenge in FL.
Notice that increasing the local step can lead to a decline in model performance due to overfitting to the local dataset.
For instance, with a local step of 1, the model accuracy is approximately 90\%, whereas with a local step of 12, it drops to around 85\%.
Through this, increasing the local step seems to still be an effective mitigation strategy, as it only reduces model accuracy by 5\% while lowering PSNR from about 18 to approximately 13.5.

\revise{\textbf{Combining multiple defense strategies.}
Beyond individual defenses, an alternative approach is to combine multiple defense strategies.
We here evaluate \sysname's performance with the combination of DP and Soteria.
In detail, we employ DP-SGD \cite{dp} in clients' local training, with $C=1$, $\delta=10^{-5}$, and noise variance calculated via $\sigma=\frac{\sqrt{2\log(1/\delta)}}{\epsilon}$ \cite{dwork2014algorithmic}.
We also implement adaptive DP \cite{fu2022adap}.
$\epsilon$ takes values of $\{ 10^5, 10^4, 10^3, 10^2 \}$.
We apply Soteria with a pruning strength parameter of 0.5 to process model updates.
Figure~\ref{imagenet_example_defense} visualizes reconstructed examples of Lung-Colon Cancer.
Overall, when $\epsilon \geq 10^3$, \sysname is still able to extract some semantic information.
Complete protection is achieved only when $\epsilon \leq 10^2$.
This observation contrasts with common literature \cite{dp} where $\epsilon=10^2$ often provides negligible privacy guarantees, primarily due to simpler datasets like MNIST or CIFAR-10.
In contrast, sophisticated datasets are more sensitive to $\epsilon$.
For example, ResNet18 with DP-SGD achieves only about 10\% accuracy at $\epsilon=71$ in ImageNet \cite{dp_imagenet}, indicating that gradients become heavily noisy.
Under such conditions, it is reasonable that \sysname struggles to extract useful private information.
Moreover, we find that the defensive capability of adaptive DP is, in fact, weaker than that of standard DP, because of adaptive DP's $\sigma$ decay mechanism, which reduces gradient noise injection over time.}

\subsection{\revise{Attack Cost}}

\begin{figure}
    \centering
    \includegraphics[width=0.8\linewidth]{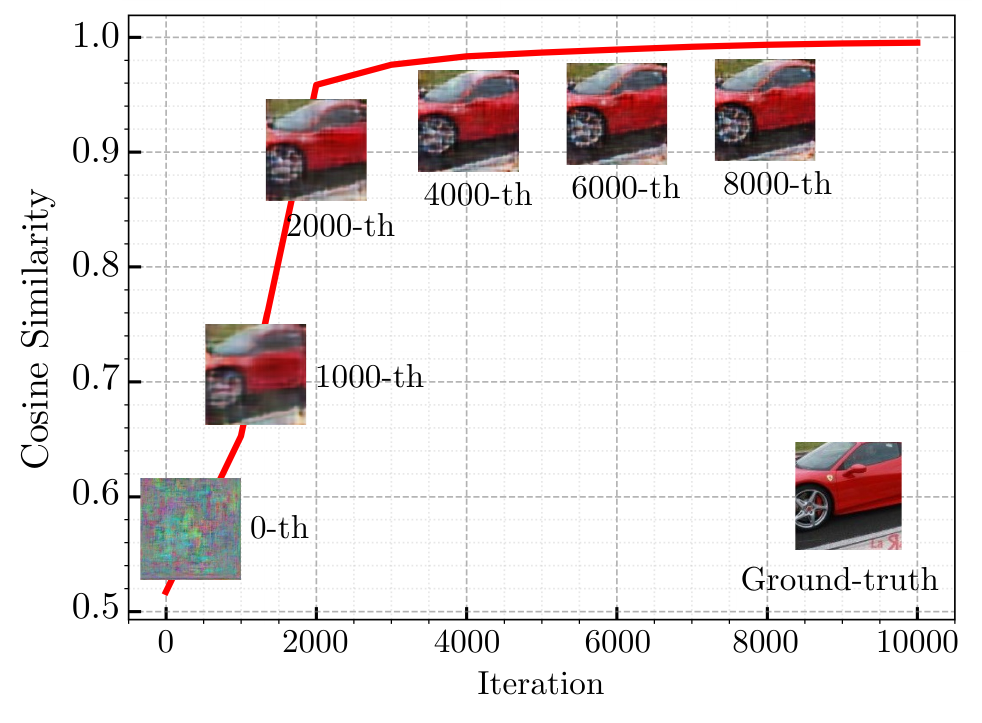}
    \caption{\revise{The gradient similarity between reconstructed images and ground-truth ones. We use ImageNet, ResNet10, and a batch size of 64.}}
    \label{fig_loss_trend}
\end{figure}

\revise{We here evaluate the computational cost of \sysname using an NVIDIA RTX 4090 GPU.
Figure \ref{fig_loss_trend} shows the convergence curve of \sysname, where we observe that approximately 1000 iterations suffice to recover the most semantic information from the ground-truth images.
Beyond 4000 iterations, the reconstructed images show only negligible perceptual changes.
For computational efficiency, \sysname requires approximately 20 minutes to complete 10000 iterations.
For 10000 iterations, GI and IG require about 5 minutes, while GGL and ROGS take about 11 minutes.
GI and IG converge within approximately 1000 iterations, whereas GGL and ROGS stabilize around 2000 iterations.
This eliminates the possibility of non-convergence in GLAs.
While \sysname indeed incurs higher computational costs compared to existing GLAs, its cost remains within an acceptable range.}

\section{Mitigation and Future Work}
\label{}

\begin{figure}[!t]
    \centering
    \includegraphics[width=0.6\linewidth]{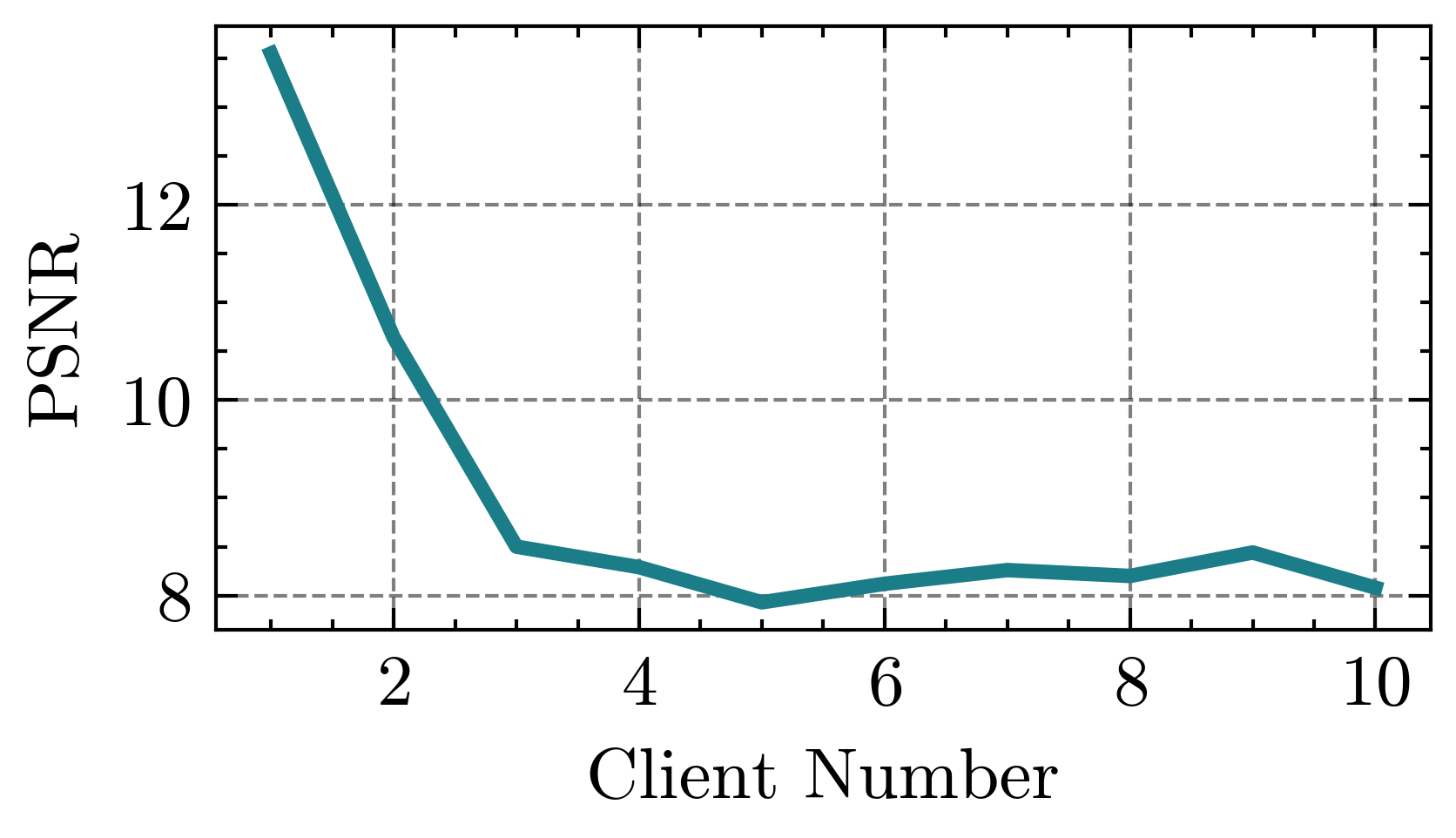}
    \caption{\revise{Performance of cryptography-based methods \cite{MPC2} against \sysname with varying participating clients.}}
    \label{fig_mpc}
\end{figure}

\begin{figure}[!t]
    \centering
    \includegraphics[width=0.6\linewidth]{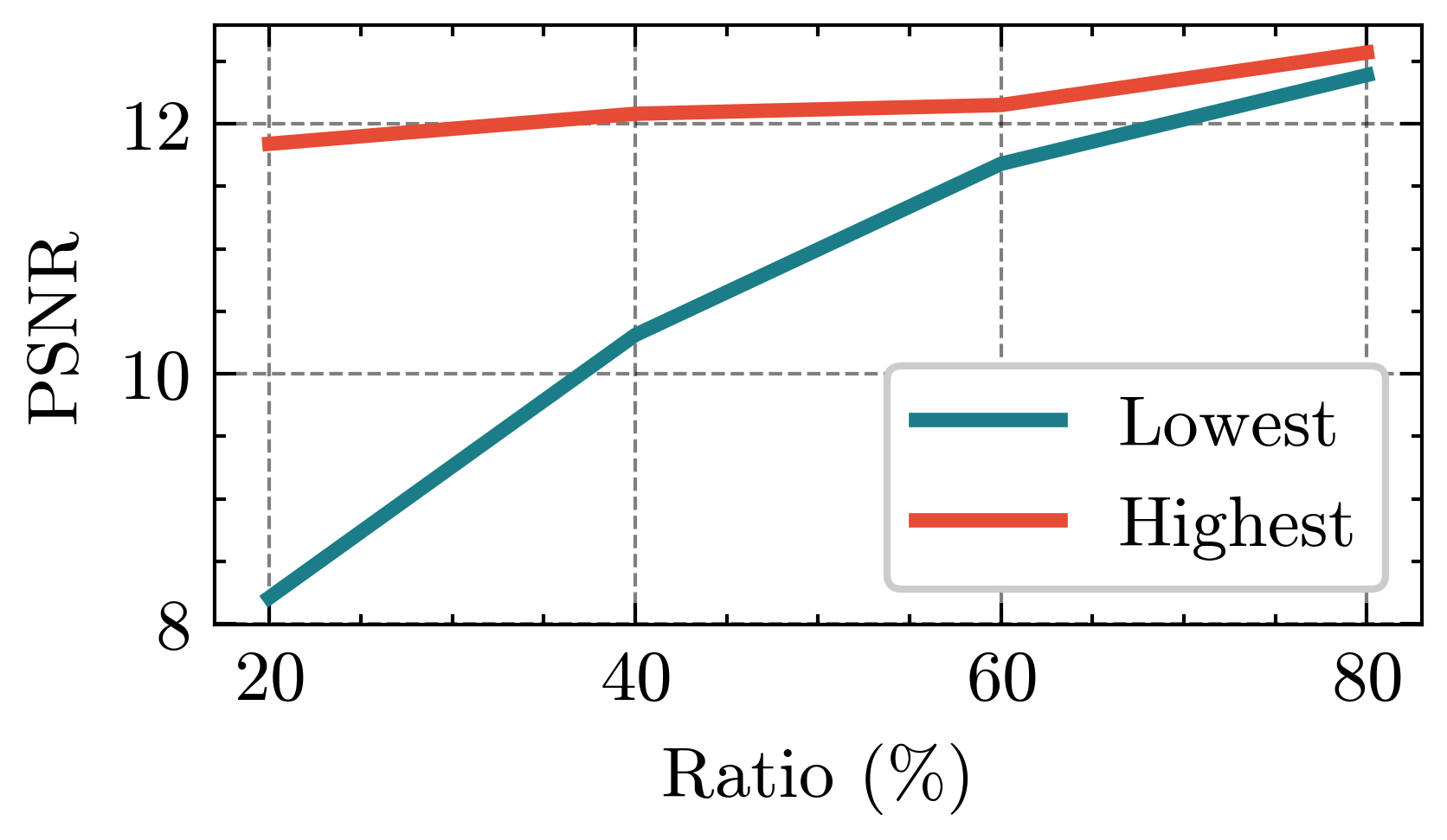}
    \caption{\revise{The performance of \sysname over two different gradient pruning strategies.}}
    \label{fig_grad_magn}
\end{figure}

\revise{Section \ref{eval_defense} shows that perturbation-based methods are insufficient to fully resist \sysname.
We also observe that increasing local steps can significantly degrade the performance of \sysname, primarily due to the amplification of the batch size.
Cryptography-based methods \cite{MPC2} exploit this by ensuring that clients only observe the plaintext of aggregated gradients, essentially scaling the batch size by the number of participating clients.
Figure \ref{fig_mpc} evaluates the performance of cryptography-based methods \cite{MPC2} against \sysname with varying participating clients per round.
Notably, \sysname's performance diminishes remarkably with just two clients per round and becomes ineffective with three clients (resulting in a batch size of 576). 
However, cryptography-based methods are often limited in practical deployment due to their complex protocol requirements and computationally intensive operations.
For a configuration of 12 local steps, a batch size of 16, and ResNet-10 on Lung-Colon-Cancer, cryptography-based methods incur substantial extra computation time: 702.03 seconds/client and 793.79 seconds/server per communication round \cite{MPC2}.
To put this computational cost into perspective, the time required for clients' local training is only 2.6 seconds.}

\revise{\textbf{What can we learn from \sysname?}
One insight in \sysname is the importance of restricting server access to gradient elements with the largest magnitudes\footnote{Intuitively, if \sysname converges, the elements with the greatest magnitude in $g$ and $g'$ will be the same. The gradient elements in $g'$ play an important role for \sysname.}.
Figure \ref{fig_grad_magn} shows \sysname's effectiveness when the server can access varying proportions of the highest and lowest magnitude gradient elements. 
When the server can only observe low-magnitude components (e.g., the bottom 20\%), \sysname achieves very low PSNR, whereas access to high-magnitude components (top 20\%) enables reconstruction performance comparable to full gradients.
This insight could guide improvements in existing methods.}

\revise{For instance, cryptography-based methods can encrypt just the top 80\% of gradient elements by magnitude, as the remaining ones are less likely to contain private information.
This selective encryption could reduce cryptographic overhead by approximately 20\%.
Another potential solution is for clients to inject noise proportionally to the magnitude of gradient elements so as to obfuscate high-magnitude elements.
Consider noise vector $\mathcal{M}$.
To counterbalance the impact of this noise on model performance, the client could send $-\mathcal{M}$ to another client, which adds it to its own model updates.
In this way, the noise vector would be canceled out in aggregation process.
However, this method requires that clients do not disclose $\mathcal{M}$ to the server.
Furthermore, we highlight that all defense methods inherently entail trade-offs among model utility, privacy preservation, computational efficiency, and threat model assumptions.
Future research should investigate these trade-offs to develop more effective defense strategies.}


\bibliographystyle{plainnat}
\bibliography{references}

\appendix
\begin{table}[!th]
\centering
\caption{The gradient distance (GD) and PSNR using different attack strategies in MNIST and CIFAR-10.}
\label{appendix_supp_revist}
\scriptsize
\begin{tabular}{@{}ccccc@{}}
\toprule
\multirow{2}{*}{Attack Strategy}            & \multicolumn{2}{c}{MNIST} & \multicolumn{2}{c}{CIFAR-10} \\ \cmidrule(l){2-5} 
                                            & GD & PSNR  & GD   & PSNR   \\ \midrule
BayesAttack                                 & 0.19              & 10.99 & 0.21                & 11.08  \\
IG                    & 0.25              & 8.70  & 0.16                & 10.32  \\ 
GI                           & 0.17              & 12.01 & 0.19                & 12.03  \\ \midrule
BayesAttack + Multiple Start                & 0.16              & 11.11 & 0.20                & 11.10  \\
IG + Multiple Start   & 0.24              & 9.01  & 0.16                & 10.33  \\ 
GI + Multiple Start          & 0.16              & 12.23 & 0.18                & 12.05  \\ \midrule
BayesAttack + Cosine Annealing              & 0.15              & 11.04 & 0.19                & 11.32  \\
IG + Cosine Annealing & 0.23              & 9.22  & 0.14                & 10.65  \\ 
GI + Cosine Annealing        & 0.14              & 12.05 & 0.16                & 12.57  \\ \bottomrule
\end{tabular}
\end{table}

\begin{figure}[!th]
    \centering
    \includegraphics[width=0.6\linewidth]{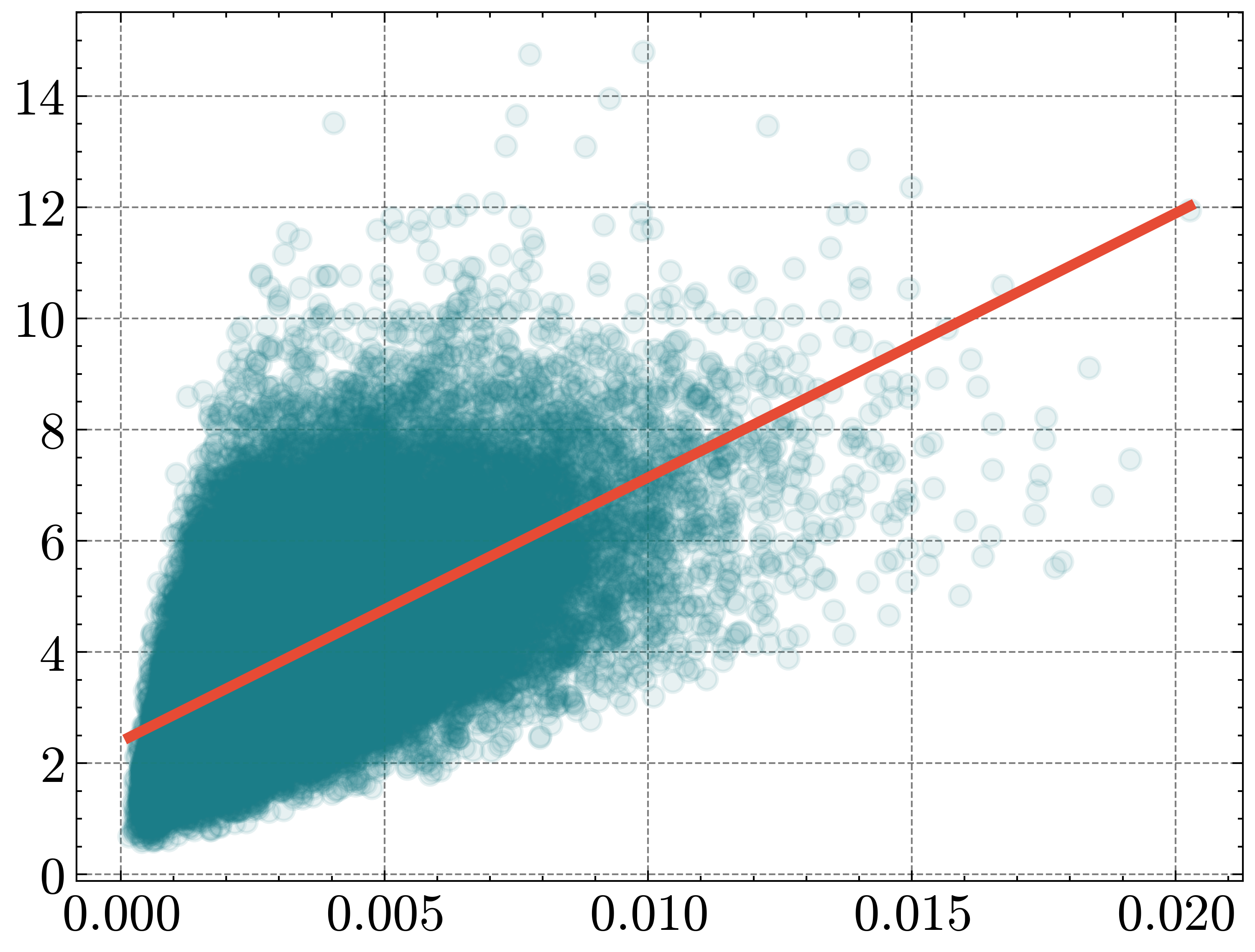}
    \caption{The x-axis represents the absolute value of the gradient elements ($|g'[i]|$), while the y-axis represents the sensitivity of gradient elements to input changes ($\sum_j |\frac{\partial g'[i]}{\partial x'[j]}|$).}
    \label{appendix_magn}
\end{figure}

\section{Supplementary Experiment for Section \ref{sec_revisit}}
\label{appendix_exp_section3}

We employ three attack methods (BayesAttack \cite{bayes_attack}, IG \cite{invertinggrad}, GI \cite{gradinversion}) on a three-layer fully connected network (Section \ref{sec_revisit}), with results in Table \ref{appendix_supp_revist}.
Note that IG uses cosine distance while others use $L_2$ distance.
The PSNR values of the reconstructed data are notably low in both MNIST and CIFAR-10 datasets, as evidenced by PSNRs around 10.

We also incorporate multiple restart \cite{pgd} and cosine annealing learning rate \cite{cos_lr} for these attacks.
Multiple restarts mitigate saddle points and local minima through repeated attacks with different seeds to avoid poor initializations.

Cosine annealing involves a sharp increase in the step size after a predetermined number of iterations (1000 in our experiments) to escape saddle points and local minima.
As reported in Table \ref{appendix_supp_revist}, although these strategies can somewhat improve attack performance, the quality of the reconstructed data remains significantly low.
Thus, the recovery of noisy data is unlikely to be mainly attributed to saddle points and local minima.

\begin{figure}[!t]
    \centering
    \subfloat{
        \includegraphics[width=0.5\linewidth]{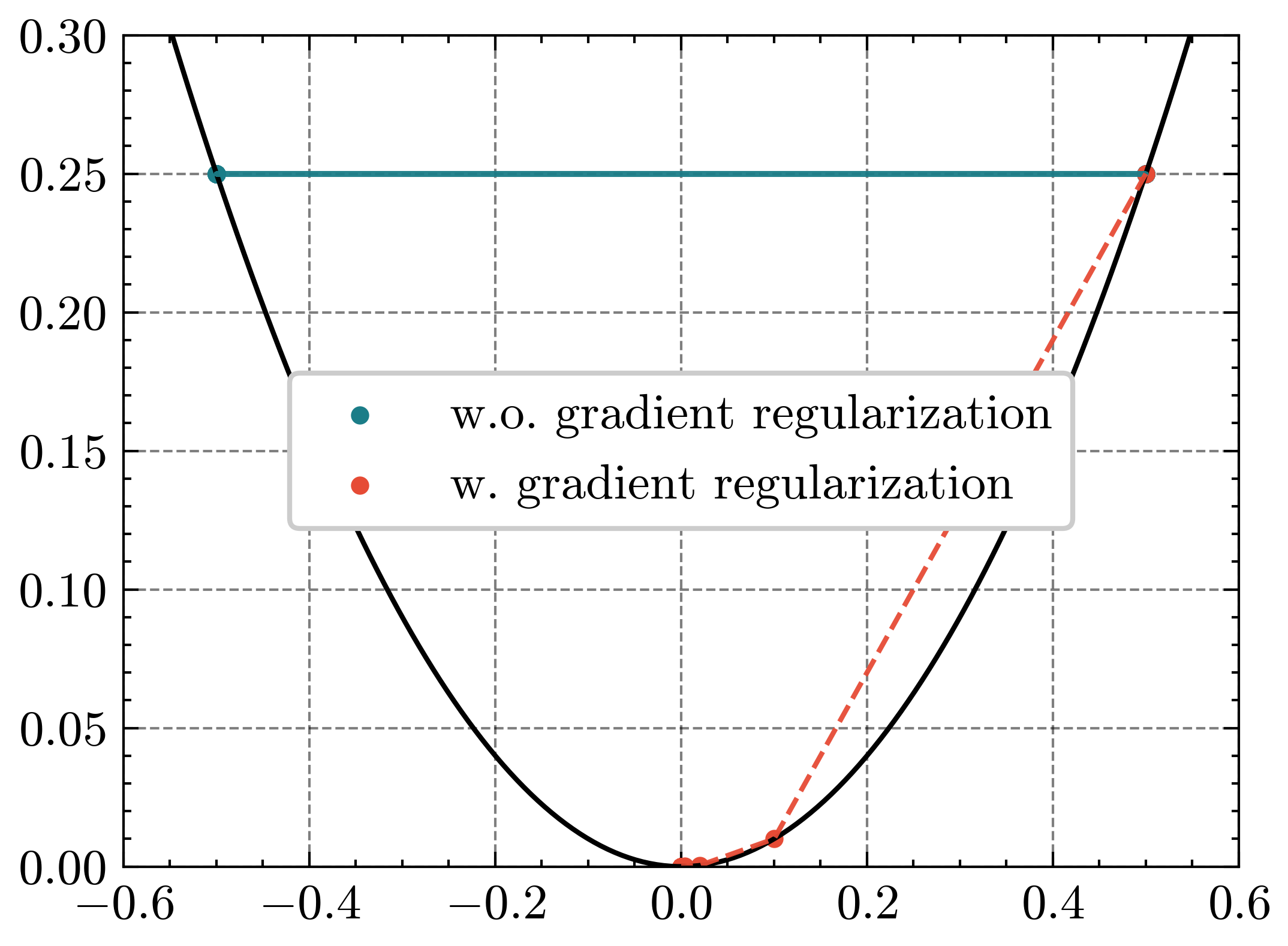}
        \label{appendix_ex_gr_1}
    }
    \subfloat{
        \includegraphics[width=0.5\linewidth]{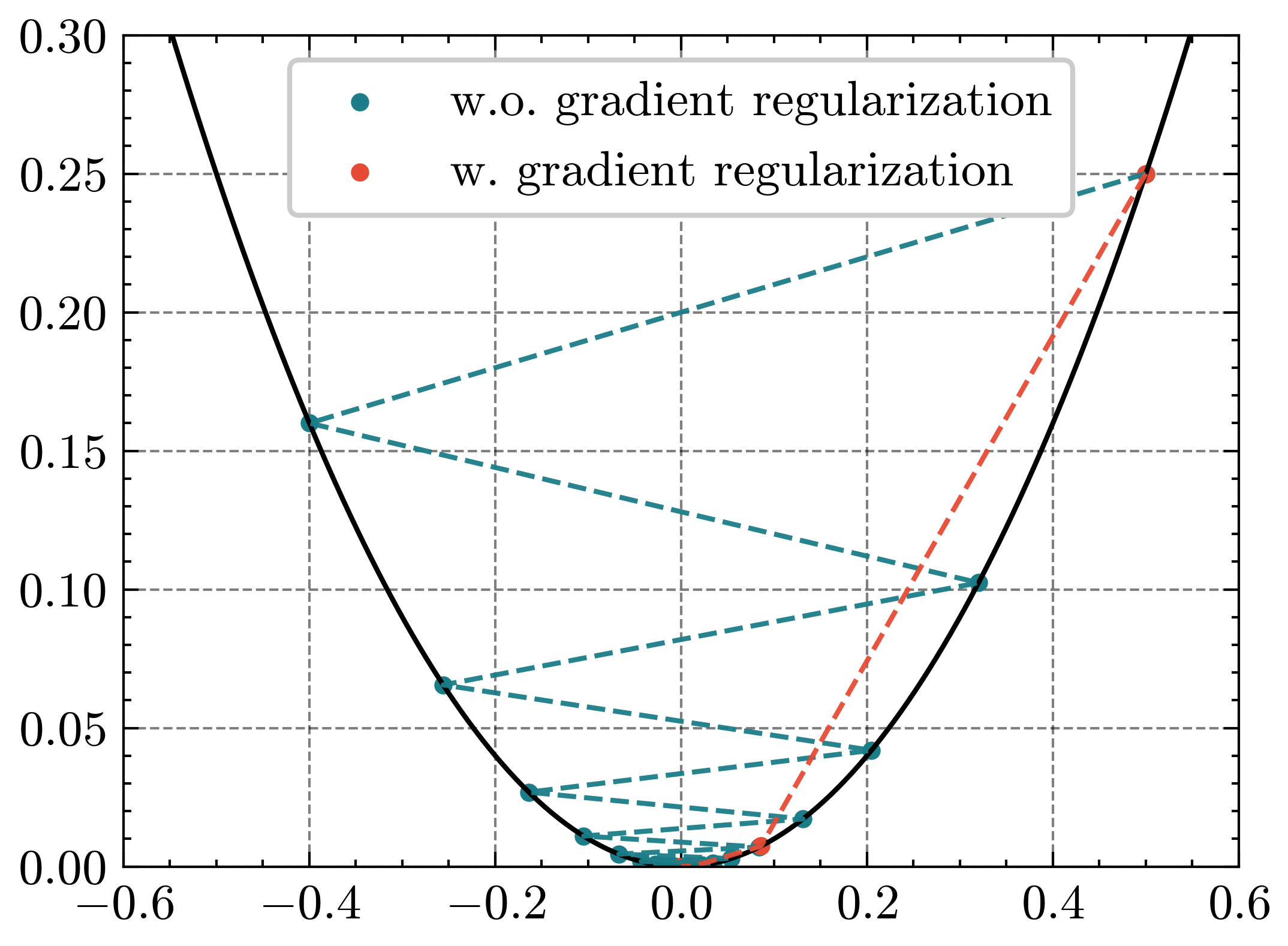}
        \label{appendix_ex_gr_2}
    }
    \caption{The search paths of the gradient descent algorithm with and without gradient regularization for minimizing $y=x^2$. The left image is with a step size of 1. The right image is with a step size of 0.9.}
    \label{appendix_grad_regular_example}
\end{figure}

\section{Gradient Estimation Error}
\label{appendix_grad_est_error}

Here, we analyze the scenario where $\hat{g} \neq \nabla_{w}\mathcal{L}(F(x, w), y)$.
We denote $\nabla_{w}\mathcal{L}(F(x, w), y)$ as $g$.
Let us suppose $\epsilon = \hat{g} - g$, with $\hat{g}$ corresponding to the gradients generated by $x^*$, i.e., $\hat{g} = \nabla_{w}\mathcal{L}(F(x^, w), y)$.
According to the mean value theorem, we have $ \hat{g} - g = \nabla^2 \mathcal{L}(F(z, w), y) \ (x^* - x) $, where z is between $x$ and $x^*$.
Consequently, according to basic theorem in algebra~\cite{convex_optimization}, if $\nabla^2 \mathcal{L}(F(z, w), y)$ is positive definite, we obtain the following inequality:
\begin{equation}
\nonumber
||\hat{g} - g|| = ||\nabla^2 \mathcal{L}(F(z, w), y) \ (x^* - x)|| \geq \sigma_{\text{min}} \ ||x^* - x||,
\end{equation}
where $\sigma_{\text{min}}$ denotes the smallest eigenvalue of $\nabla^2 \mathcal{L}(F(z, w), y)$.
This indicates that when the estimation error is $\epsilon$, the distance between ground-truth data associated with $\hat{g}$ and $g$ is bounded by $\frac{||\epsilon||}{||\nabla^2 f(z)||}$.
Notice that in practice, if the model fits the dataset well, the Hessian matrix ($\nabla^2 \mathcal{L}(F(z, w), y)$) should indeed be positive definite.
Moreover, a similar analytical framework can be applied to the estimation errors associated with $y$ and we will not elaborate on that further here.

\section{Empirical Validation about Section \ref{sec_less_is_better}}
\label{appendix_partial}

Here, we empirically demonstrate that gradient elements with larger magnitudes are more sensitive to changes in input data.
We employ ResNet10 and CIFAR-10.
Specifically, we input 100 samples into the model and calculate the gradients of the final linear layer ($g'[i]$).
We then calculate the absolute sum of each gradient element with respect to the input data ($\sum_j |\frac{\partial g'[i]}{\partial x'[j]}|$) and visualize the results in Figure~\ref{appendix_magn}.
As can be seen, there is a clear trend: as the magnitude of the gradient elements increases, so does their sensitivity to input changes.
The Pearson correlation coefficient is $0.7372 \pm 0.0566$.
In statistics, correlation coefficient above 0.7 is generally considered significant.
Thus, gradient elements with larger magnitudes indeed respond more sensitively to variations in the input data.

\section{Example for Gradient Regularization}
\label{appendix_gr_example}

To clarify the role of gradient regularization (Equation \ref{eq_15}), we illustrate its effect on minimizing $y=x^2$ using gradient descent with and without regularization.
Starting at $x=0.5$, as illustrated on the left side of Figure \ref{appendix_grad_regular_example}, a fixed step size of 1 without gradient regularization causes oscillation between $0.5$ and $-0.5$. Even with a reduced learning rate 0.9, oscillations persist, slowly converging to $x=0$.  

Equation \ref{eq_15} combines gradients at the current point $x$ and a subsequent point along the descent path.
Intuitively, if the gradient at the forward point opposes that at the current point, it indicates that the step size is too large.
Then, the gradient at the forward point will shrink the gradient at the current point $x$, which indirectly reduces the step size.
This behavior is illustrated in Figure \ref{appendix_grad_regular_example}, where we set $\lambda'$ to 0.3.
The gradient at $x=0.5$ is 1, while the gradient at the next point $x=-0.5$ is $-1$.
Consequently, the resulting mixed gradient is $(1-\lambda') \times 1 + \lambda' \times  (-1)=0.4$.
This demonstrates how gradient regularization facilitates the optimization process.

\section{Supplementary Investigation Result}
\label{appendix_investigation}

\begin{table*}[!ht]
\centering
\caption{A summary of critical factors influencing the performance of GLAs.}
\label{appendix_facor_investigation}
\scriptsize
\begin{tabular}{ccp{11cm}}
\toprule
Factor                 & Literature & Explanation \\ \midrule
Batch Size             & \cite{sok_grad_leakage,eval_frame1,eval_frame2,dlg,gradinversion,grad_gene3,LGLA}      &  A larger batch size introduces more optimization variables, which complicates the data reconstruction process.            \\ 
Model                  & \cite{invertinggrad,LGLA,guardian}      &  Intuitively, a larger model should impose more constraints on the gradient matching problem, potentially enhancing the reconstruction quality. However, this increased complexity also leads to heightened nonlinearity of the model. Compared to linear problems, non-linear problems are notoriously more difficult to solve. Early GLAs present poor performance on large models, likely due to their inability to effectively address the gradient matching problem associated with large models.            \\
Dataset                & \cite{sok_grad_leakage,eval_frame2,invertinggrad,grad_gene1,grad_gene2,grad_gene3,LGLA}      &    Higher-resolution images suggest more optimization variables, thereby exacerbating reconstruction difficulty.          \\
Attack Stage           & \cite{sok_grad_leakage,bayes_attack}      &  GLAs tend to yield better attack results during the early stages of training, which can be attributed to the model's higher sensitivity at that time. This is elaborated in Section \ref{sec_introduction}.            \\
Initialization Method  & \cite{invertinggrad,out}      &   Different weight initialization methods can affect the model's sensitivity to data, thereby influencing the reconstruction difficulty. For instance, wide uniform initialization can amplify the weight intensity, causing minor data variations to lead to significant changes in the model's output. This increased sensitivity can facilitate data reconstruction.           \\
Auxiliary Information & \cite{sok_grad_leakage,eval_frame1,grad_gene1,out}      &  Apparently, auxiliary information gives the server more advantages in solving the gradient matching problem.            \\
Local Step             & \cite{sok_grad_leakage,eval_frame2,out}      &When the batch size is fixed, an increase in local steps leads to uploaded model updates being derived from more data, which complicates the data reconstruction. Besides, an increase in local steps also can introduce greater estimation errors on the server side, adding another layer of difficulty to the data reconstruction.         \\ \bottomrule
\end{tabular}
\end{table*}

\begin{table}[!ht]
\caption{The default batch sizes of introductory examples in various open-source libraries.}
\label{appendix_default_batch}
\centering
\scriptsize
\begin{tabular}{@{}cc@{}}
\toprule
Library              & Batch Size \\ \midrule
NVIDIA FLARE         & 4-32       \\
FLOWER               & 32-64      \\
Substra              & 32         \\
PySyft               & 64         \\
OpenFL               & 32         \\
TensorFlow Federated & 20         \\ \bottomrule
\end{tabular}
\end{table}

Upon review of existing research related to GLAs, we identify eight key factors that significantly impact the performance of these attacks and summarize these factors in Table \ref{appendix_facor_investigation}.
These factors are either explicitly mentioned or shown to have a substantial impact on the effectiveness of GLAs.
To collect common values for eight factors in production environments, we reviewed extensive FL literature on FL and open-source FL libraries.
The attack stage is server-determined, while auxiliary information is largely scenario-driven, such as whether the specified scenario can collect substantial data similar to that of the clients.
We focus on the remaining seven factors, including batch size, model, dataset, initialization method, auxiliary information, local step, and defense.
\begin{itemize}[leftmargin=*,topsep=1pt]
    \item \textbf{Batch size.} Some survey papers \cite{common_batch} indicate that the common batch size ranges from 16 to 64.
    Our examination of application cases published in journals such as Nature, Science, and BMC over the past several years reveals that many studies use vague expressions like "specific/common parameters" without providing precise values.
    A few papers specify batch sizes of 16 \cite{batch_of_16}, 32 \cite{batch_of_32,batch_of_32_2}, 60\footnote{\url{https://github.com/gkaissis/PriMIA/blob/master/configs/torch/pneumonia-conv.ini}}, and 64 \cite{batch_of_64}.
    We conjecture that these studies predominantly adopt the default recommended values from existing frameworks.
    Introductory examples in common industrial libraries also typically use batch sizes between 20 and 64, as summarized in Table \ref{appendix_default_batch}.
    Therefore, we consider the range of 16 to 64 as standard practice.
    \item \textbf{Model.} \citet{model_usage} compiled the studies utilizing FL in the medical domain. Among the 47 papers they reviewed, 35 employed CNN architectures, primarily ResNet, DenseNet, and MobileNet, while 4 utilized transformer models. This highlights that CNNs remain the dominant architecture in practical applications. This is likely due to their higher computational efficiency and lower data requirements compared to transformer models \cite{vit}, making them better suited for data-scarce medical environments.
    \item \textbf{Dataset.} As noted in most literature \cite{model_usage, init_method, batch_of_16, batch_of_64}, real-world datasets typically have higher resolutions compared to commonly used benchmark datasets, which are often limited to a resolution of $32 \times 32$.
    \item \textbf{Initialization method.} \citet{init_method} summarized the weight initialization approaches of existing FL application papers in the medical domain.
    Among the 89 papers they reviewed, 68 explicitly discussed approaches for weight initialization.
    Among 89 studies, 45/89 used random initialization, 16/89 pre-determined parameters, and 7/89 pre-trained weights.
    Pre-determined parameters can vary significantly across tasks.
    As such, we focus on random initialization and pre-trained weights.
    \item \textbf{Local step.} Most papers vaguely describe local steps. Common values include 1 step or 1 local epoch \cite{localstep1,localstep2,batch_of_32}.
    \item \textbf{Defense.} \citet{init_method} reviewed 89 FL application papers and found that most did not actively implement defense methods. Specifically, six papers applied DP, five of which were based on Gaussian mechanism. It seems that practitioners in real-world scenarios generally have strong confidence in FL's privacy protection capabilities. Although the use of privacy-enhancing methods is not yet widespread in practice, we still evaluate the effectiveness of \sysname against defense methods.
\end{itemize}

\begin{figure*}[!t]
    \centering
    \subfloat[ImageNet]{
        \includegraphics[width=0.32\linewidth]{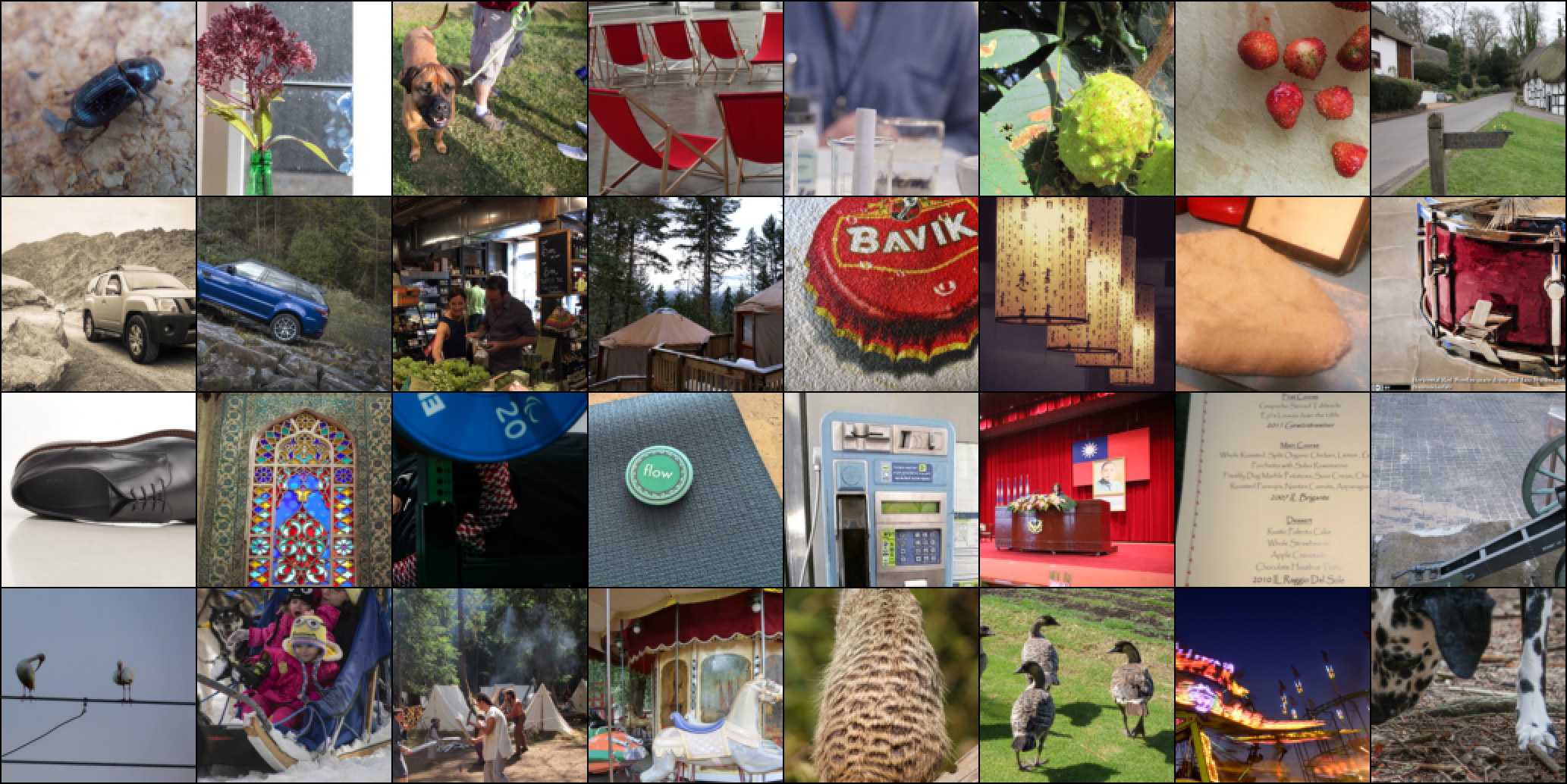}
    }
    \subfloat[HAM10000]{
        \includegraphics[width=0.32\linewidth]{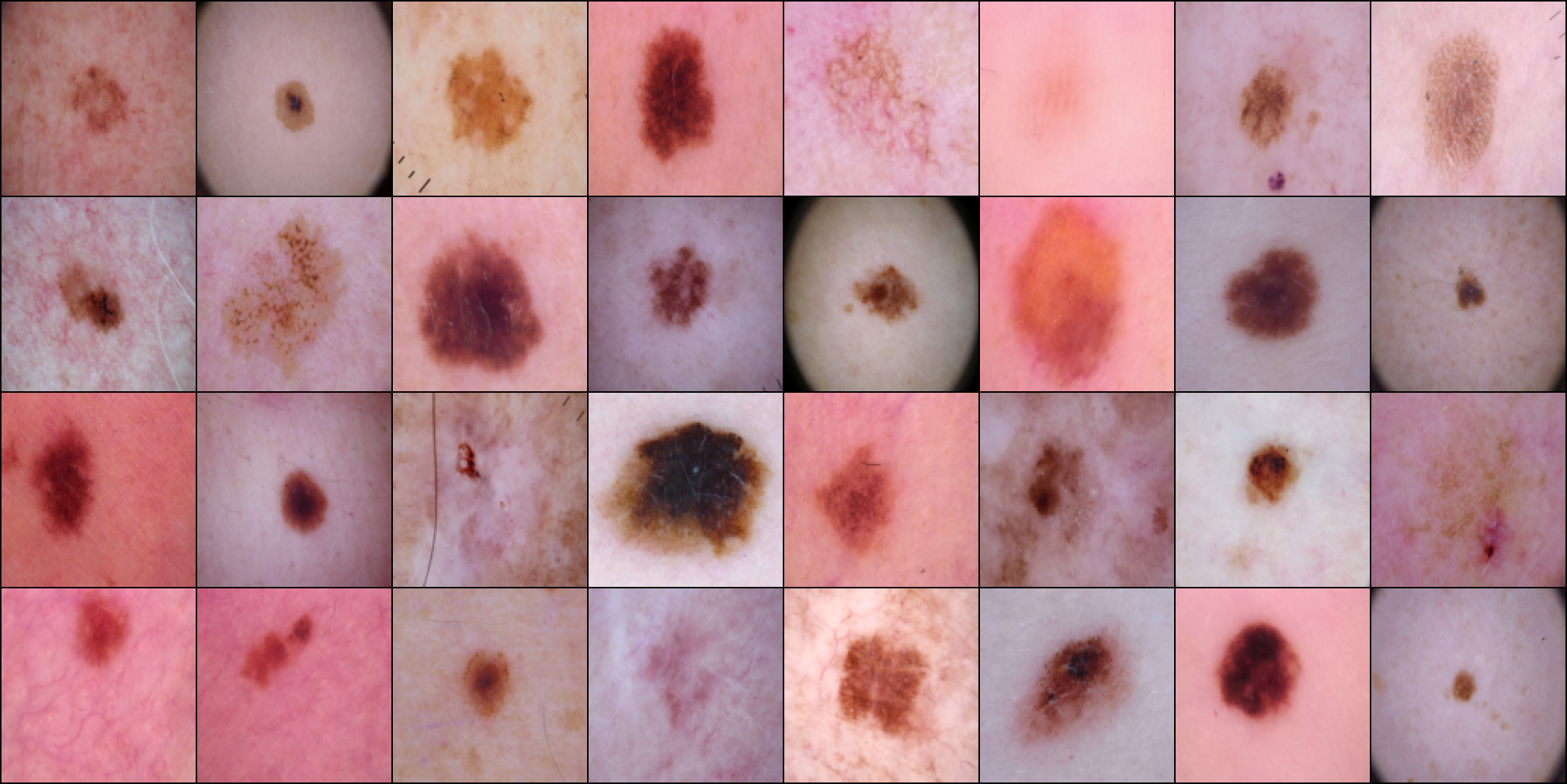}
    }
    \subfloat[Lung-Colon Cancer]{
        \includegraphics[width=0.32\linewidth]{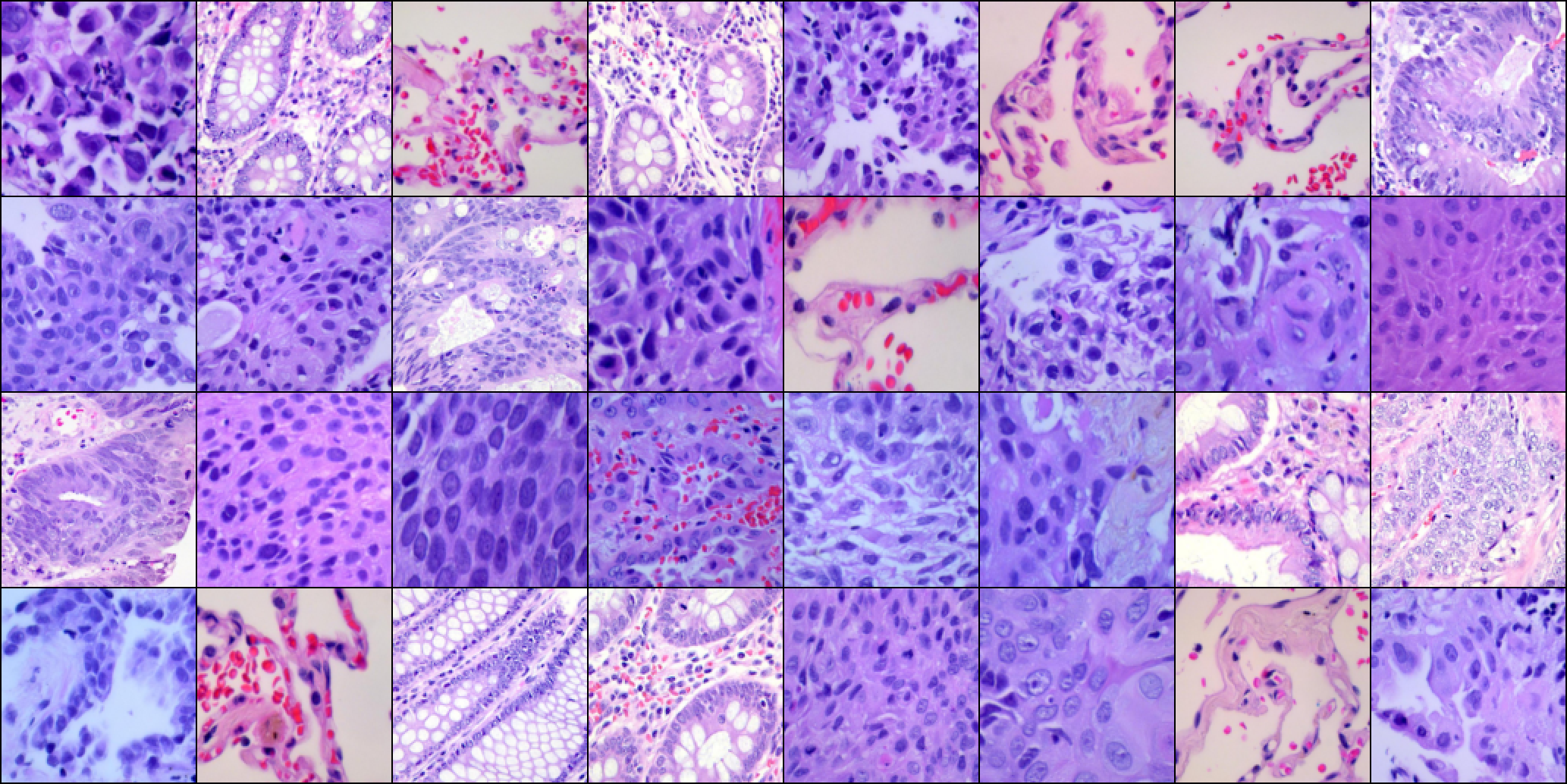}
    }
    \caption{The ground-truth batch of Figure \ref{visual_complex_dataset}.
    }
    \label{appendix_real_imgs}
\end{figure*}

\begin{figure}[!t]
    \centering
    \includegraphics[width=0.7\linewidth]{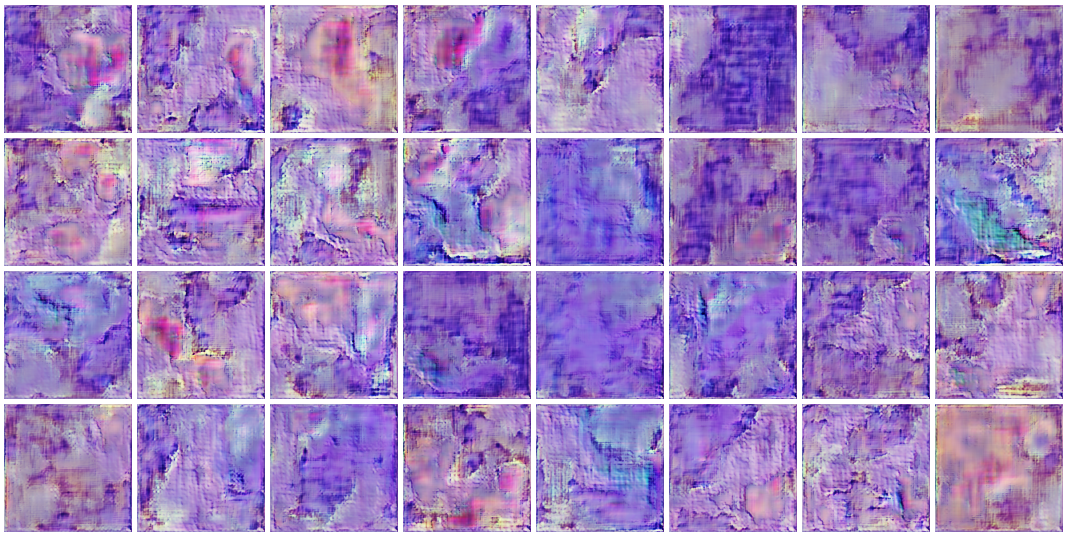}
    \caption{\revise{ROGS's recovered batch in Lung-Colon Cancer.}}
    \label{ROGS_recovered}
\end{figure}

\section{Label Inference, Sensitivity Analysis, and Ablation Study}
\label{appendix_sens_ablation}

We first discuss label inference.
We then study the impact of different hyperparameters on the performance of \sysname.
By default, we use a batch size of 32 and CIFAR-10.

\begin{table}[!ht]
\caption{\revise{Label inference accuracy and performance metrics across different batch sizes in ImageNet.}}
\label{tab_label_infer}
\centering
\scriptsize
\begin{tabular}{@{}cccc@{}}
\toprule
Batch Size & \begin{tabular}[c]{@{}c@{}}Label Inference\\ Accuracy\end{tabular} & \begin{tabular}[c]{@{}c@{}}PSNR\\ (with Label Inference)\end{tabular} & \begin{tabular}[c]{@{}c@{}}PSNR\\ (w.o. Label Inference)\end{tabular} \\ \midrule
32         & 100\%                                                               & 19.07                                                                 & 17.94                                                                 \\
64         & 96.88\%                                                             & 19.01                                                                 & 17.50                                                                  \\
128        & 92.19\%                                                             & 18.40                                                                  & 16.45                                                                 \\ \bottomrule
\end{tabular}
\end{table}

\revise{
\textbf{The impact of label inference.}
Table~\ref{tab_label_infer} reports the accuracy of label inference and the \sysname's performance with and without label inference across different batch sizes, over 1280 ImageNet samples.
We see label inference generally achieves high accuracy and enhances reconstruction quality.
}

\begin{table}[!h]
\caption{The impact of $R$.}
\label{table_impact_r}
\centering
\scriptsize
\begin{tabular}{@{}ccccccc@{}}
\toprule
$R$ & 15\%  & 30\%  & 50\%  & 65\%  & 80\%  & 100\% \\ \midrule
PSNR           & 18.85 & 20.34 & 21.33 & 20.76 & 19.80 & 15.06 \\ \bottomrule
\end{tabular}
\end{table}

\textbf{The impact of matching ratio $R$.}
We here incrementally increase the value of $R$ from 15 to 100 to observe the performance change of \sysname, as illustrated in Table \ref{table_impact_r}.
Looking at the results we see a performance trend that initially rises and peaks at $R=50$, followed by a subsequent decline.
This trend can be intuitively understood as a result of balancing the number of matched gradient elements.
Matching too few gradient elements might lead to the omission of many useful elements, whereas matching too many can lead to an overly small $\mu$ value.

\begin{table}[!h]
\caption{The impact of $\lambda'$.}
\label{table_impact_lambda}
\centering
\scriptsize
\begin{tabular}{@{}cccccc@{}}
\toprule
$\lambda'$ & 0.1   & 0.3   & 0.5   & 0.7   & 0.9   \\ \midrule
PSNR   & 18.85 & 20.34 & 21.33 & 20.76 & 19.80 \\ \bottomrule
\end{tabular}
\end{table}

\textbf{The impact of blend factor $\lambda'$.}
A higher $\lambda'$ heightens emphasis on gradient regularization, thereby reducing the value of $L$.
Nonetheless, an excessively high $\lambda'$ might disproportionately prioritize optimizing the gradient regularization term, potentially neglecting the gradient matching.
As anticipated, Table \ref{table_impact_lambda} illustrates this rising and then falling trend.

\begin{table}[!h]
\caption{The impact of $\alpha$.}
\label{table_impact_alpha}
\centering
\scriptsize
\begin{tabular}{@{}ccccccc@{}}
\toprule
$\alpha$ & 0     & $ 10^{-5}$  & $10^{-4}$  & $10^{-3}$  & $ 10^{-2}$  & $ 10^{-1}$  \\ \midrule
PSNR  & 21.05 & 21.33 & 21.99 & 21.01 & 17.79 & 14.31 \\ \bottomrule
\end{tabular}
\end{table}

\begin{table}[!h]
\caption{The impact of $\beta$.}
\label{table_impact_beta}
\centering
\scriptsize
\begin{tabular}{@{}cccccc@{}}
\toprule
$\beta$ & 0     & 0.0001 & 0.001 & 0.01  & 0.1   \\ \midrule
PSNR & 20.84 & 21.33  & 21.80 & 21.17 & 20.20 \\ \bottomrule
\end{tabular}
\end{table}

\textbf{The impact of $\alpha$ and $\beta$.}
Table \ref{table_impact_alpha} and Table \ref{table_impact_beta} present the impact of varying $\alpha$ and $\beta$ on \sysname's performance.
Both $\alpha$ and $\beta$ exhibit a rising and subsequently falling pattern regarding attack performance.
Specifically, an overly large $\alpha$ leads to overly smooth constructions, where all pixel values converge to the same number.
Likewise, a high $\beta$ causes the recovered images to have activation values close to zero, thus reconstructing most pixel values to zero.

\begin{table}[!h]
\caption{The impact of loss function.}
\label{table_impact_loss}
\centering
\scriptsize
\begin{tabular}{@{}cccc@{}}
\toprule
Loss Function & w.o. cos & w.o. $L_1$ & cos \& $L_1$   \\ \midrule
PSNR & 20.04  &  19.86   & 21.33 \\ \bottomrule
\end{tabular}
\end{table}
\textbf{Ablation study on loss function.}
We further assess the performance of \sysname when removing either the $L_1$ distance or cosine distance from Equation (\ref{eq_16}).
As reported in Table \ref{table_impact_loss}, omitting either distance measure results in performance degradation, showing the necessity of both components in \sysname's effectiveness.

\section{Supplementary Theory}
\label{appendix_supp_theory}

\begin{thm}
\label{appendix_theory1}
    Removing the minimum values from a set $S=\{a_1, \cdots, a_n\}$ and then calculating the average of the remaining values will result in an increase in the average.
\end{thm}

\begin{proof}
Without loss of generality, let $a_n$ be the smallest element in the set $S$. 
The average of the original $S$ is given by $\text{Average}(S) = \frac{1}{n} \sum_{i=1}^{n} a_i.$
After removing the smallest element $a_n$, the new set is defined as $S' = S \setminus \{ a_n \}.$
The average of the new set $S'$ can be expressed as $\text{Average}(S') = \frac{1}{n-1} ( \sum_{i=1}^{n} a_i - a_n )$.
We demonstrate that $\text{Average}(S') \geq \text{Average}(S).$
This inequality can be rewritten as 
$\frac{1}{n-1} ( \sum_{i=1}^{n} a_i - a_n) \geq \frac{1}{n} \sum_{i=1}^{n} a_i.$
Cross-multiplying yields:
$$n ( \sum_{i=1}^{n} a_i - a_n ) \geq (n-1) \sum_{i=1}^{n} a_i.$$
Expanding and rearranging gives:
$$n \sum_{i=1}^{n} a_i - n a_n \geq n \sum_{i=1}^{n} a_i -  \sum_{i=1}^{n} a_i.$$
Simplifying leads to $n a_n \leq \sum_{i=1}^{n} a_i.$
Since $a_n$ is the minimum value in $S$, it follows that $n a_n \leq \sum_{i=1}^{n} a_i$ holds true.
We can iteratively remove subsequent minimum elements and apply the same reasoning.
Each removal of the smallest element results in an increased average of the resulting set, thereby confirming the validity of Theorem \ref{thm2}.
\end{proof}

\end{document}